%% file: iclr2027_conference.tex
\documentclass{article} 
\usepackage{iclr2027_conference,times}
\usepackage{iftex}
\ifXeTeX
  \usepackage{fontspec}
  \IfFontExistsTF{TeX Gyre Termes}
    {\setmainfont{TeX Gyre Termes}}
    {\setmainfont{NimbusRoman}[
       Path=/usr/share/fonts/opentype/urw-base35/,
       Extension=.otf,
       UprightFont=*-Regular,
       BoldFont=*-Bold,
       ItalicFont=*-Italic,
       BoldItalicFont=*-BoldItalic]}
\fi

\usepackage{hyperref}
\hypersetup{hidelinks}
\usepackage{url}
\usepackage{amssymb,mathtools}
\usepackage{booktabs}
\usepackage{tabularx}
\usepackage{enumitem}
\usepackage{graphicx}
\usepackage{wrapfig}
\usepackage{algorithm}
\usepackage{algorithmic}


\title{TRACE: Typed Reference-Aware Conversational Editing for Humanoid Motion Diffusion}

\author{%
\begin{minipage}[t]{\dimexpr\textwidth-2\tabcolsep\relax}
\centering
\normalsize\bfseries
Zejian Kang\textsuperscript{1,2,*}\quad
Yifu Luo\textsuperscript{3,*}\quad
Boyang Shen\textsuperscript{4,*}\quad
Jingxuan Li\textsuperscript{5}\\[3pt]
Zekai Chen\textsuperscript{6}\quad
Zekun Shi\textsuperscript{7}\quad
Yongxue Xu\textsuperscript{8}\quad
Zhenyu Hou\textsuperscript{9}\\[3pt]
Bingqian Wu\textsuperscript{2}\quad
Shaolei Zhang\textsuperscript{10}\quad
Wentao Yang\textsuperscript{1,2}\quad
Xiangru Huang\textsuperscript{2,\textdagger}\\[9pt]
\normalfont\small
\textsuperscript{1}Zhejiang University\quad
\textsuperscript{2}Westlake University\\[2pt]
\textsuperscript{3}Technical University of Munich\\[2pt]
\textsuperscript{4}Huazhong University of Science and Technology\\[2pt]
\textsuperscript{5}University of Science and Technology Beijing\\[2pt]
\textsuperscript{6}Beijing Institute of Technology\quad
\textsuperscript{7}Beihang University\\[2pt]
\textsuperscript{8}Sun Yat-sen University\quad
\textsuperscript{9}Nanjing University\\[2pt]
\textsuperscript{10}Renmin University of China\\[7pt]
\textsuperscript{*}Equal contribution.\quad
\textsuperscript{\textdagger}Corresponding author.\\[2pt]
\href{mailto:huangxiangru@westlake.edu.cn}{huangxiangru@westlake.edu.cn}
\end{minipage}%
}
\hypersetup{
  pdftitle={TRACE: Typed Reference-Aware Conversational Editing for Humanoid Motion Diffusion},
  pdfauthor={Zejian Kang, Yifu Luo, Boyang Shen, Jingxuan Li, Zekai Chen, Zekun Shi, Yongxue Xu, Zhenyu Hou, Bingqian Wu, Shaolei Zhang, Wentao Yang, Xiangru Huang}
}

\iclrfinalcopy
\begin{document}

\maketitle
\pagestyle{plain}
\thispagestyle{plain}

\begin{abstract}
Language-conditioned humanoid motion generation enables users to create
and refine robot movements through natural language. The central challenge
is to connect an underspecified request to concrete motion structure that
can subsequently be revised relative to an accepted reference. We present
TRACE, a generation and editing framework for the Unitree G1 humanoid
robot built around typed, measurable kinematic quantities. Our insight is
to ground generation conditions in paired training motions and editing
goals in the current motion. A language model learns to predict support,
root, and hand events from motion-derived supervision; their time-aligned
encodings complement the original text in a diffusion generator.
Gravity-referenced states and trajectory--contact objectives improve
kinematic consistency. For editing, typed requests define body--time scopes
and source-relative goals. A separate reference-conditioned diffusion editor
uses these goals and revised events to realize changes, while sampling-time
projection preserves protected state coordinates. Quantitative evaluation
and human assessment show improved semantic alignment and motion quality
over the compared methods. Multi-turn
editing evaluations demonstrate more reliable execution of successive
relative requests under shared preservation constraints.
Code is available at \url{https://github.com/Multimodal-Perception-Action/TRACE}.
\end{abstract}

\input{sec/1_intro}

\input{sec/2_related_work}
\input{sec/3_method}
\input{sec/4_exp}

\input{sec/editing_experiments}
\input{sec/5_conclusion}
\bibliography{iclr2027_conference}
\bibliographystyle{iclr2027_conference}

\appendix
\input{sec/appendix}

\end{document}

%% file: sec/1_intro.tex
\section{Introduction}
\label{sec:introduction}

\begin{figure*}[t]
 \centering
 \includegraphics[width=0.96\textwidth]{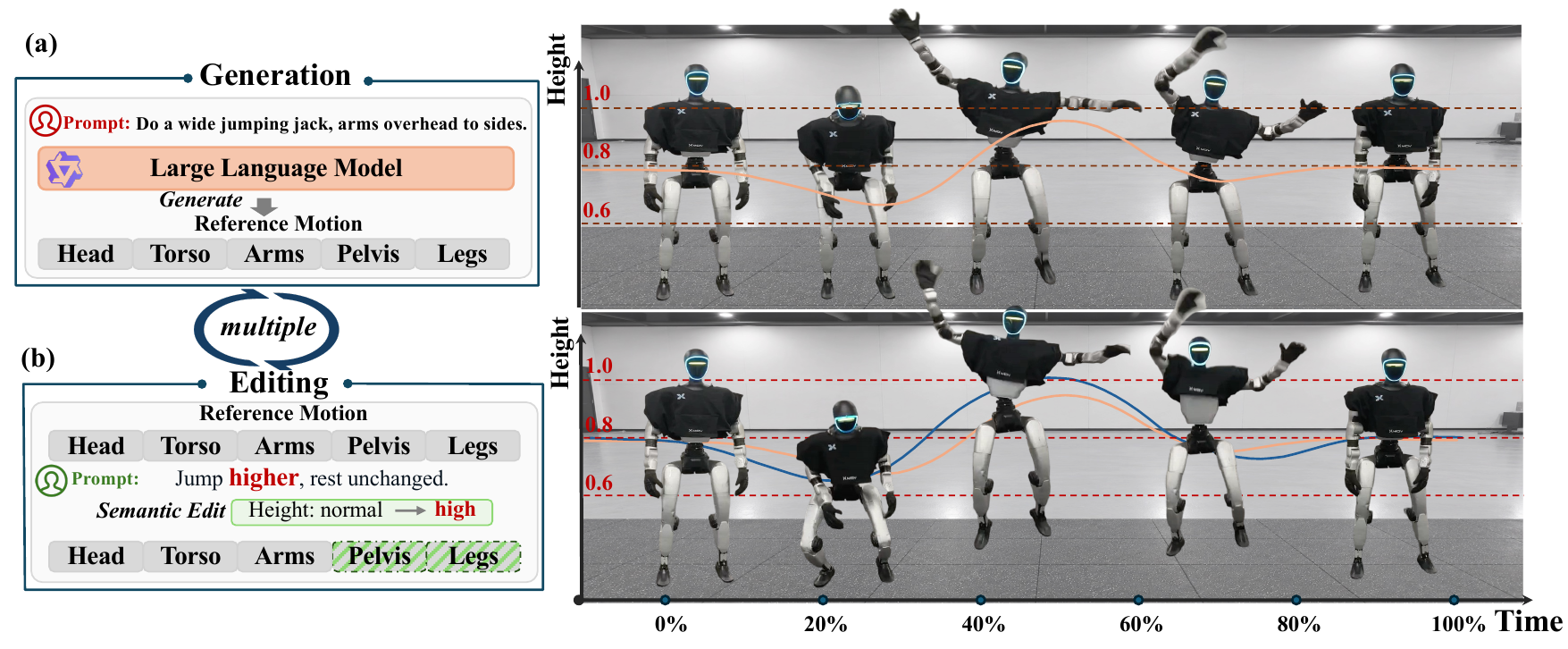}
 \caption{\textbf{From text generation to reference-aware refinement.}
(a) The request \lq\lq{}Do a wide jumping jack, arms overhead to sides'' produces
an initial motion reference. (b) \lq\lq{}Jump higher, rest unchanged'' specifies
a relative height change; the schematic highlights the pelvis and legs
involved in the revision. The robot sequence shows a deeper preparatory
crouch and a higher jump. Orange and blue height profiles compare the
initial and edited motions over normalized time.}
 \label{fig:teaser}
\end{figure*}

Language-conditioned humanoid motion generation maps a natural-language
instruction to a sequence of root poses and joint configurations. It lets
users specify behavior such as walking, dancing, or jumping without
authoring each pose. The resulting trajectory can serve as an animation
or a reference for a downstream robot controller. In interactive use,
generation also provides a starting point for refinement: after seeing a
motion, a user may ask for a higher jump or a different hand position while
expecting the accepted parts of the sequence to remain recognizable.

Existing text-to-motion models learn mappings from captions to continuous
trajectories or discrete motion codes \citep{mdm,t2mgpt,motiongpt,momask}.
Structured approaches expose body-part or temporal conditions
\citep{finemogen,motionscript}, and language-model planning decomposes
instructions into detailed descriptions \citep{motionr1}. These intermediate
conditions provide a way to specify motion structure, but a caption alone
rarely determines exactly when a foot supports the body or how high a hand
rises. A plan inferred solely from that caption can describe a valid
realization different from its paired training motion. Grounding the plan
in measurable properties of the demonstration gives supervision a direct
connection to the trajectory the generator learns to reconstruct.

Editing makes this connection to a particular motion equally important.
Source-conditioned methods learn revisions from language and an existing
sequence \citep{motionfix,simmotionedit,cime}. A relative request such as
\lq\lq{}raise the hand'' must be interpreted against the current pose and
trajectory, while the rest of the motion supplies a reference to preserve.
An interactive system therefore needs both an explicit description of the
requested change and a mechanism that restricts its effect to the selected
body region and time interval.

The \emph{key challenge} is to turn underspecified language into a coherent
motion reference whose intended properties can be revised selectively.
Figure~\ref{fig:teaser} illustrates this requirement. An initial instruction
produces a jumping jack; the follow-up \lq\lq{}Jump higher, rest unchanged'' refers
to that particular motion. The edited example crouches deeper and jumps
higher, requiring coordinated changes in preparation and take-off while
retaining the overall action. A useful intermediate description should
therefore expose quantities such as support and height over time.
Our insight is that these measurable quantities can connect both stages:
the paired demonstration determines the event targets used to train
generation, and the current reference determines the baseline and protected
coordinates for editing.

We introduce \emph{Typed Reference-Aware Conversational Editing} (TRACE)
for the Unitree G1 humanoid robot. \emph{Typed} refers to explicit categories
for support, root, and hand events and for requested changes.
\emph{Reference-aware} means that the current motion defines both the baseline
for a relative change and the coordinates to preserve. For generation, we
extract event targets from training motions through forward kinematics and
supervise an LLM to predict duration and event intervals from text.
The predicted Plan is converted into numerical conditions aligned with
motion patches. A diffusion Transformer combines these conditions with
the original caption, retaining language information beyond the compact
event vocabulary. A gravity-referenced state directly represents height
and tilt, while integrated-trajectory and support-selective contact
objectives constrain motion consistency.

For editing, a language request is parsed into a typed change and compiled
into a body--time scope and a source-relative goal. A separately trained
diffusion editor receives the current motion, its revised events, and the
compiled goal. At every sampling step, reference projection restores
protected coordinates, allowing the selected region to change while
retaining the rest of the encoded motion. Each output supplies the reference
for the next request. Generation and editing thus share a motion
representation and a kinematic vocabulary, with separate models for their
respective conditioning tasks.

Experiments on BONES-SEED~\citep{bonesseed} evaluate generation, kinematic
consistency, and successive request execution. TRACE leads the compared
methods in retrieval, MM-Dist, FID, and human ratings
(Section~\ref{sec:overall_performance}). Matched controls establish the
value of event conditioning and motion-derived auxiliary supervision,
while representation and objective ablations measure their effects on
trajectory and contact errors (Section~\ref{sec:ablation_studies}).
For editing, hand-height requests provide measurable source-relative goals;
TRACE achieves higher all-turn success than the adapted learned editors
under shared scopes and reference projection
(Section~\ref{sec:multiround_editing}). Our contributions are:
\begin{enumerate}
    \item \textbf{Motion-grounded typed-event conditioning.}
    We derive confidence-weighted support, root, and hand event targets
    from training trajectories and learn a text-to-Plan predictor.
    Patch-aligned numerical events complement raw-caption features
    in continuous G1 motion generation. Matched controls demonstrate
    gains from event conditioning and motion-derived auxiliary targets.

    \item \textbf{Gravity-referenced motion diffusion.}
    We directly represent absolute height and gravity-relative tilt,
    and combine integrated-trajectory objectives with support-selective
    forward-kinematics penalties to reduce accumulated trajectory
    errors and improve foot--ground consistency. Representation and
    objective ablations quantify these effects through kinematic errors.

    \item \textbf{Typed, source-relative motion editing.}
    We compile supported editing requests into body--time scopes and
    source-relative goals, and combine them with source-motion features
    and revised events in a separately trained editor.
    Reference projection preserves protected state coordinates throughout
    sampling. Shared-scope comparisons demonstrate more reliable successive
    hand-height changes to an evolving motion reference.
\end{enumerate}

%% file: sec/2_related_work.tex
\section{Related Work}
\label{sec:related_work}

\paragraph{Motion-grounded planning and conditioning.}
Text-to-motion models generate discrete codes
\citep{t2mgpt,motiongpt,momask} or continuous trajectories \citep{mdm}.
Structured representations add finer control: MotionScript converts motion
into detailed language \citep{motionscript}, and FineMoGen models body-part
and temporal composition \citep{finemogen}. Motion-R1 synthesizes decomposed
reasoning plans from descriptions and refines generation with RL
\citep{motionr1}. For supervised motion generation, the intermediate
conditions also need to agree with the specific demonstration paired with
the caption. TRACE addresses this alignment through confidence-weighted
support, root, and hand events extracted from that training motion. A
text-conditioned planner predicts these events at inference, and their
numerical sequence supplements raw-caption features with event content
and temporal alignment.

\paragraph{Kinematic consistency and humanoid generation.}
MDM incorporates geometric and foot-contact objectives \citep{mdm}.
PhysDiff adds simulator-based motion projection during denoising
\citep{physdiff}, while CLoSD closes the loop between a diffusion planner
and a tracking controller \citep{closd}. Language-conditioned humanoid
systems connect generation to robot motion or control
\citep{harmon,langwbc,roboghost,bibo}. OMG~\citep{omg} conditions
diffusion-based humanoid motion generation on language, audio, and human
reference motions. Imitation and transfer methods
address downstream tracking \citep{humanplus,omnih2o,asap}.
These methods address motion plausibility and downstream execution at
different stages. TRACE targets upstream reference quality through
gravity-referenced height
and tilt, integrated trajectory objectives, and support-selective FK
contact losses. 

\paragraph{Typed source-relative and reference-preserving editing.}
MDM demonstrates temporal and body-part inpainting \citep{mdm}, and
PriorMDM composes diffusion priors for joint and trajectory control
\citep{motionprior}. FineMoGen also uses LLMs for fine-grained editing
\citep{finemogen}. MotionFix introduces source--instruction--target pairs
and the source-conditioned TMED editor \citep{motionfix};
SimMotionEdit adds motion-similarity supervision \citep{simmotionedit},
and CIME studies spatial preservation and non-uniform temporal mapping
\citep{cime}. TRACE connects source conditioning and preservation to
explicit relative requests: a typed
relative change is compiled into a kinematic body--time scope and coupled
to source-relative goals, revised source events, and reference projection.

%% file: sec/3_method.tex
\section{Method}
\label{sec:method}

\begin{figure*}[t]
 \centering
 \includegraphics[width=0.98\textwidth]{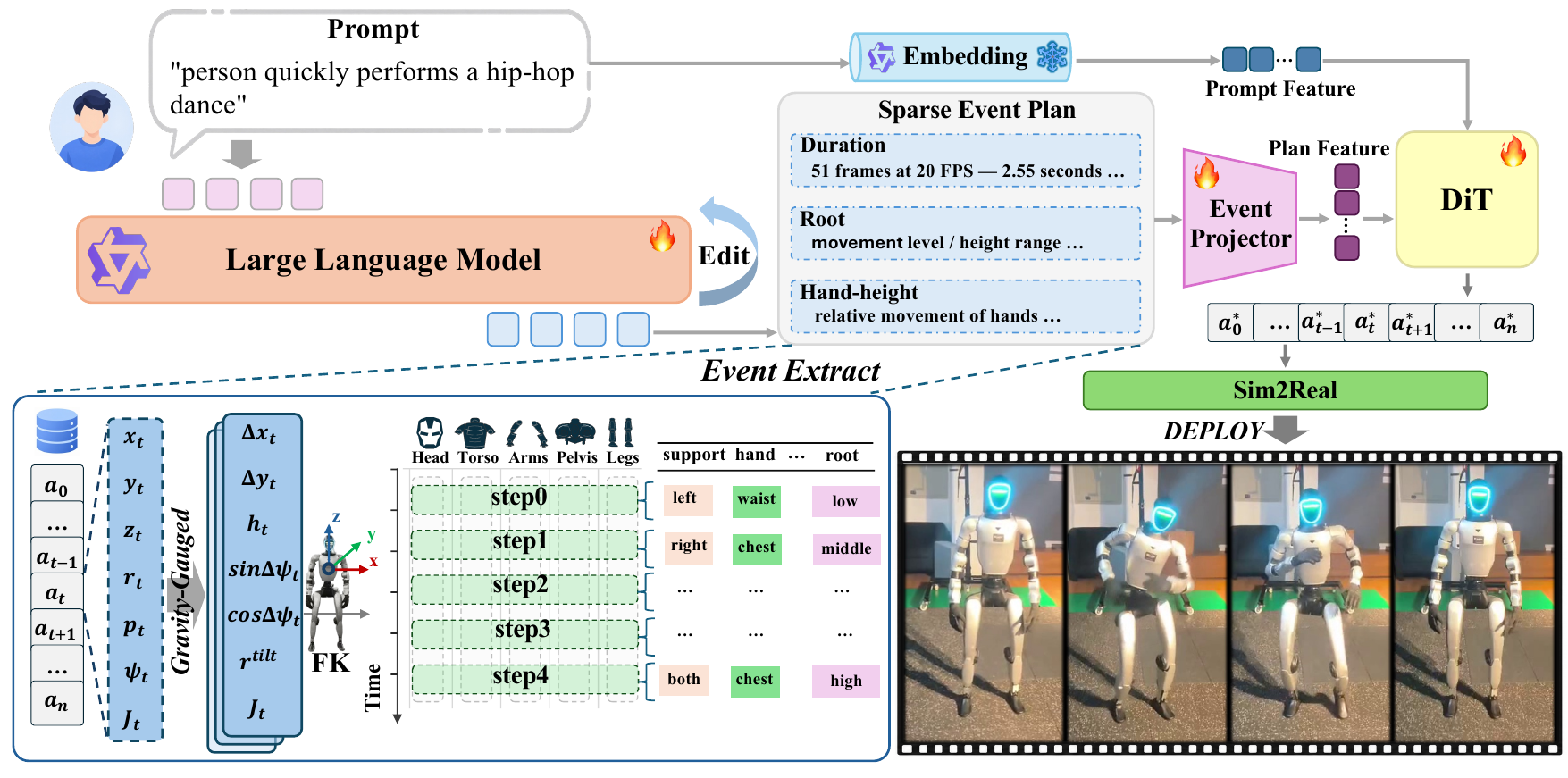}
 \caption{\textbf{TRACE with a motion-grounded sparse-event plan.}
 The upper path maps a text instruction to continuous G1 motion. A frozen
 text encoder supplies prompt features, while an LLM predicts duration and
 typed support, root, and hand events. The event projector maps their
 numerical, time-aligned encoding to plan features that condition the
 diffusion Transformer (DiT) alongside the prompt. The lower-left path
 constructs training supervision: gravity-gauged states and forward
 kinematics (FK) provide measurable signals for event extraction, linked
 to the plan by the dashed path. Test-time plans are predicted from text.
 Snowflake and flame symbols denote frozen and trainable modules in their
 respective training stages; the trained LLM is frozen for diffusion
 training. Generated references can be passed to a downstream controller
 for robot execution, illustrated at lower right. The edit loop denotes
 follow-up requests; Section~\ref{sec:interactive_state} describes the
 separate reference-conditioned editor.}
 \label{fig:method_overview}
\end{figure*}

\subsection{Overview}
\label{sec:overview}

Typed event plans structure generation; typed edit requests specify how an
accepted motion should change. Reference-conditioned denoising realizes
the change within its compiled scope, while projection preserves protected
coordinates. TRACE uses separate generator and editor weights;
Figure~\ref{fig:method_overview} depicts the generation pathway.

For an instruction $u$, a frozen text encoder gives $e=f(u)$, while a
supervised LLM jointly predicts duration and typed event intervals.
Deterministic rasterization converts its structured output into the plan:
\begin{equation}
 P=(N,C)=\operatorname{Rasterize}(\operatorname{LLM}_{\phi}(u)),
 \quad C\in[0,1]^{L\times16},\quad L=\lceil N/4\rceil.
 \label{eq:online_graph}
\end{equation}
The LLM predicts support, root, and hand events and is frozen during
diffusion training and inference. We denote its rasterized events at
length $N$ by $h(u,N)$.
A conditional diffusion Transformer generates continuous G1 states from
$(e,P)$.

\subsection{Motion-Grounded Sparse Planning and Conditioning}
\label{sec:typed_program}

We construct a compact event vocabulary from measurable properties of
the paired training motion. These motion-derived targets provide
realization-aligned supervision for the language-to-plan predictor.
At inference, predicted events supply explicit support, root, and hand
structure, while the original caption retains semantics outside this
vocabulary.

\paragraph{Motion-derived supervision.}
For each clean training motion, deterministic FK and root reconstruction
produce sole heights/speeds, root vertical motion and height, and palm
heights relative to the pelvis. Fixed thresholds convert these signals
into typed events. Each four-frame patch carries two independent support
probabilities and four categorical distributions:
\begin{equation}
 c_\ell=[c_\ell^{L\mathrm{foot}},c_\ell^{R\mathrm{foot}},
         c_\ell^{\mathrm{root\text{-}v}},
         c_\ell^{\mathrm{root\text{-}h}},
         c_\ell^{L\mathrm{hand}},c_\ell^{R\mathrm{hand}}],
 \qquad c_\ell\in[0,1]^{16}.
 \label{eq:program_slots}
\end{equation}
These semantic groups form a compact vocabulary, with categorical entries
summing to one within each valid group. \emph{Sparse} refers to this
vocabulary; every valid motion patch receives a condition.

\paragraph{Plan prediction and training conditions.}
Threshold ambiguity is encoded by confidence $W$.
Motion-derived pseudo-targets supervise the LLM through serialized typed
plans using only training motions; $Y$ denotes their numerical encoding.
To reduce dependence on perfect annotations at deployment, diffusion
training mixes target and predicted conditions while retaining raw text.
Uncertain targets fall back to predicted events:
\begin{equation}
 C^{\mathrm{mix}}=
 \begin{cases}
 W\odot Y+(1-W)\odot h(u,N), & \text{with probability }1/2,\\
 h(u,N), & \text{otherwise}.
 \end{cases}
 \label{eq:plan_mixture}
\end{equation}
The choice applies jointly to all event groups for a sample. Training uses
the recorded length; inference uses the caption-predicted length and
predicted events only. Appendix~\ref{sec:plan_annotation_rules} specifies
the kinematic extraction rules and an actual training example.

\paragraph{Time-aligned conditioning.}
The raw caption is encoded once. A learned numerical event projection
and a separate raw-text residual branch use patch progress features to
align conditions to the denoiser. Both supplement the raw-text backbone.
At diffusion timestep $k$, the residual condition is scaled by
$\alpha(k)=\min(1,k/[0.25(K-1)])$, attenuating it near clean motion.
Matched controls retain the same branch capacity, so removing plan means
zeroing its event inputs while preserving all trainable parameters.

\subsection{Gravity-Gauged and Contact-Consistent Diffusion}
\label{sec:generation}
\label{sec:training}

The plan specifies motion structure, but accurate local poses alone do not
ensure a coherent trajectory. Small errors in integrated root motion can
accumulate, and plausible poses can still move a supporting foot. We
therefore represent height and tilt relative to gravity and supervise
integrated geometry and foot contact explicitly. This separates quantities
that require an absolute reference from those that describe displacement.

\paragraph{Gravity-gauged states and kinematic consistency.}
We predict absolute height and gravity-relative tilt directly, integrating
only planar displacement and yaw. The per-frame state is
\begin{equation}
 x_t=[\Delta p_t^x,\Delta p_t^y,h_t,
       \sin\Delta\psi_t,\cos\Delta\psi_t,\mathbf r_t^{\mathrm{tilt}},j_t],
 \label{eq:gauge_root}
\end{equation}
Here $\mathbf r_t^{\mathrm{tilt}}$ concatenates the first two columns of
$H_t^\top R_t$, where $R_t$ is the body-to-world root rotation and $H_t$
is its horizontal heading rotation; $j_t$ contains 29 joint angles.
The six tilt coordinates are orthonormalized and composed with integrated
yaw to recover the root rotation, up to the canonical initial heading
(Appendix~\ref{sec:gravity_gauge_codec}). This yields a 40-dimensional
motion representation. Plan support probabilities serve only as denoiser
conditions.

\paragraph{Grouped training objective.}
The generator predicts clean normalized states under a cosine noise
schedule. We organize its objective into three weighted groups:
\begin{equation}
 \mathcal L_{\mathrm{gen}}=
 \mathcal L_{\mathrm{motion}}+
 \mathcal L_{\mathrm{geometry}}+
 \mathcal L_{\mathrm{contact}}.
 \label{eq:generation_objective}
\end{equation}
\emph{Motion reconstruction} supervises states, velocities, and accelerations.
\emph{Geometry consistency} constrains reconstructed position, cumulative
yaw, body-up direction, and the yaw-pair norm. \emph{Contact consistency} uses differentiable
FK to penalize support-foot sliding, drift, and height error, plus penetration
on all valid frames. Clean training motion selects support masks; predicted
plan probabilities are conditions, not substitutes for these masks.
Each group includes its component weights, detailed in
Appendix~\ref{sec:contact_implementation}.

\subsection{Automatic Masking and Reference-Aware Motion Editing}
\label{sec:interactive_state}
\label{sec:local_diffusion}

Editing requires both locating the requested change and preserving the
remaining motion. We combine automatic body--time masking, source-relative
goals, and reference projection to address these complementary requirements.

\paragraph{Automatic scope localization.}
At turn $t$, $X_t=(M_t,u_t)$ contains the source and request;
$N_t$ is the source length. A supervised LLM, frozen at inference,
predicts $A_t=\delta_t=\pi_\phi(u_t,N_t,s(M_t))$: an attribute,
effector, signed change, and interval. The source-only summary $s(M_t)$
contains FK hand heights relative to the pelvis. A deterministic compiler
constructs $(S_t,D_t)=g(A_t,u_t,N_t)$. For the body--time mask $S_t$, it
expands the selected effector $e_t$ to its kinematic chain
$\mathcal K(e_t)$ and resolves interval $[a_t,b_t)$ against $N_t$:
\begin{equation}
 S_t(f,j)=\mathbf 1[j\in\mathcal K(e_t)]\,
          \mathbf 1[a_t\leq f<b_t].
 \label{eq:automatic_scope}
\end{equation}
For hand-height edits, $\mathcal K$ contains seven arm-joint coordinates;
its complement remains protected, including that arm outside the interval.
Explicit time expressions are compiled from language; invalid requests
leave the mask closed. Localization uses only the source and request.
The grammar is detailed in Appendix~\ref{sec:interaction_mdp_details}.

\paragraph{Source-relative motion revision.}
For controlled hand-height evaluation, $D_t\in\mathbb R^{N_t\times2}$
encodes signed wrist-height offsets with a sine-squared temporal envelope.
The editor conditions on source-motion features, the scope, these offsets,
and revised source events. The event sequence is extracted deterministically
from the current source motion, with the requested hand-height groups updated
according to $D_t$. Original and synthetic edit pairs supervise local motion
fidelity, relative goals, and event consistency.
The wrist-offset protocol evaluates hand edits; Figure~\ref{fig:teaser}
provides a separate qualitative jump refinement. Appendix~\ref{sec:interaction_mdp_details}
defines request compilation, and Appendix~\ref{sec:edit_backend_validation}
details the editor and its training.

\paragraph{Reference-preserving transitions.}
The conditional editor defines a stochastic state transition:
\begin{equation}
 M_{t+1}\sim\mathcal T_\eta(\cdot\mid X_t,A_t)
 =p_\eta(\cdot\mid f(u_t),M_t,S_t,D_t).
 \label{eq:interaction_factorization}
\end{equation}
Revised source events are deterministic functions of the source motion and
compiled offsets; this dependence is implicit in the compact conditioning
notation above.
The next state is $X_{t+1}=(M_{t+1},u_{t+1})$, with an external utterance;
the Markov formulation assumes self-contained requests. After each DDIM
step $\widetilde x_{k-1}$, the binary mask $S$ preserves protected
normalized coordinates by reference projection:
\begin{equation}
 x_{k-1}=S\odot\widetilde x_{k-1}
       +(1-S)\odot q_{k-1}(\operatorname{Enc}(M_t)).
 \label{eq:masked_sampling}
\end{equation}
The reference is noised to the current sampling level; the final step
copies protected encoded state coordinates exactly. In the hand-height
protocol, the root and nonselected joint chains remain fixed. For
root-changing edits, this coordinate guarantee alone does not ensure
world-space preservation. Mask localization and editing under shared
scopes are evaluated separately.

%% file: sec/4_exp.tex
\section{Motion Generation Experiments}
\label{sec:experiments}

We evaluate generation quality (Section~\ref{sec:overall_performance})
and test Plan conditioning and kinematic design with matched controls
(Section~\ref{sec:ablation_studies}), using the setup in
Section~\ref{sec:experimental_setup}.

\subsection{Experimental Setup}
\label{sec:experimental_setup}

\paragraph{Data.}
We use the processed BONES-SEED test split of 13,302 text--motion pairs.
Each UID contributes one complete generated motion, including reassembly
of packed segments where required. Processing, split integrity, and
reassembly are detailed in Appendix~\ref{sec:bones_seed_processing}.

\paragraph{Baselines.}
We distinguish two groups. \emph{Official G1 checkpoints} comprise
TextOP~\citep{textop}, TeXedo~\citep{texedo}, and ECHO~\citep{echo}.
\emph{Human-motion methods adapted to G1} comprise
MDM~\citep{mdm}, MoMask~\citep{momask}, and MotionGPT3~\citep{motiongpt3},
retrained on the processed BONES-SEED/G1 split with heading-local 38D outputs.
TRACE is trained on the processed split and uses Plans predicted by the LLM
from text, without target-motion conditions. All methods are evaluated on
the same test set; training and inference configurations are documented in
Appendix~\ref{sec:experiment_details}.

\paragraph{Metrics.}
All rows use the same frozen G1 text--motion evaluator for 32-candidate
R-Precision, MM-Dist, FID, and Diversity. Diversity is a distributional
diagnostic, not a ranked objective. Physical errors are assessed only in
matched component controls. Table captions identify the uncertainty
measure; Appendix~\ref{sec:metric_definitions} gives definitions,
repetitions, and aggregation.

\paragraph{Implementation.}
TRACE generates 40D G1 states at 20 FPS with four-frame patches.
The diffusion Transformer has eight layers, width 448,
and eight heads. Frozen Qwen3-Embedding-4B supplies caption features;
a Qwen3-4B-Instruct adapter jointly predicts duration and all event groups. The main-table
generator continues a 32k-step raw-text model for 8k steps with the
16-event branch and frozen LLM predictions. Test inference uses predicted
conditions, EMA, DDIM-50, and CFG 2.5.
Appendix~\ref{sec:experiment_details} records the full training and
sampling configuration and batch sizes.

\subsection{Generation Performance Analysis}
\label{sec:overall_performance}

\begin{table*}[t]
\vspace{-5mm}
\centering
\caption{BONES-SEED comparison: automatic metrics on test motions
(R@k in percent) and a user study on 100 test prompts (scores 1--3).
The upper block uses official checkpoints; the lower block contains
G1-adapted baselines and TRACE trained on the processed BONES-SEED split.
Cells show means above and uncertainty below where reported: evaluator-resampling 95\% CI for R-Precision and SD across 16 participant means for the study. Automatic metrics and user-study scores follow
the protocols defined in Appendix~\ref{sec:metric_definitions}.}
\label{tab:trace_main}
\footnotesize
\setlength{\tabcolsep}{1.5pt}
\renewcommand{\arraystretch}{1.12}
\newcommand{\maincell}[2]{\shortstack{$#1$\\[-1pt]{\scriptsize $#2$}}}
\resizebox{\linewidth}{!}{%
\begin{tabular}{lccccccccc}
\toprule
& \multicolumn{6}{c}{Automatic test-set metrics} &
  \multicolumn{3}{c}{User study} \\
\cmidrule(lr){2-7}\cmidrule(lr){8-10}
Method & R@1$\uparrow$ & R@2$\uparrow$ & R@3$\uparrow$
& MM-Dist$\downarrow$ & FID$\downarrow$ & Diversity
& Quality$\uparrow$ & Text match$\uparrow$ & Overall$\uparrow$ \\
\midrule

\input{tables/main_semantic_only_rows_20260919.tex} 
\bottomrule
\end{tabular}%
}
\end{table*}

TRACE achieves the best retrieval, MM-Dist, and FID
results in Table~\ref{tab:trace_main}, indicating strong semantic
alignment and distributional fidelity of the complete system.
The user study provides complementary evidence of perceptual quality
and text alignment.

Figure~\ref{fig:in_the_wild_look_left_right} compares TRACE with
MotionGPT3 and MDM. The selected test examples qualitatively illustrate
closer semantic alignment through clearer dance gestures and
turn-then-walk ordering. Table~\ref{tab:trace_main} evaluates the complete
system; the matched controls in Section~\ref{sec:ablation_studies} assess
the contribution of motion-grounded event conditioning.

\input{fig/in_the_wild_look_left_right}

\input{sec/contribution_ablations}

%% file: tables/main_semantic_only_rows_20260919.tex
TextOP~\citep{textop} & \maincell{10.54}{\pm 0.07} & \maincell{19.02}{\pm 0.09} & \maincell{25.18}{\pm 0.11} & \maincell{1.2950}{} & \maincell{0.3985}{} & \maincell{1.2437}{} & \maincell{1.331}{\pm0.199} & \maincell{1.281}{\pm0.152} & \maincell{1.306}{\pm0.154} \\
TeXedo~\citep{texedo} & \maincell{10.78}{\pm 0.10} & \maincell{17.80}{\pm 0.09} & \maincell{22.60}{\pm 0.07} & \maincell{1.3150}{} & \maincell{0.4060}{} & \maincell{1.2505}{} & \maincell{1.275}{\pm0.191} & \maincell{1.106}{\pm0.106} & \maincell{1.191}{\pm0.125} \\
ECHO~\citep{echo} & \maincell{24.08}{\pm 0.12} & \maincell{36.90}{\pm 0.13} & \maincell{44.05}{\pm 0.13} & \maincell{1.2008}{} & \maincell{0.2577}{} & \maincell{1.3288}{} & \maincell{1.469}{\pm0.218} & \maincell{1.469}{\pm0.196} & \maincell{1.469}{\pm0.195} \\
\midrule
MoMask~\citep{momask} & \maincell{31.57}{\pm 0.08} & \maincell{41.88}{\pm 0.10} & \maincell{46.80}{\pm 0.07} & \maincell{1.1801}{} & \maincell{0.3057}{} & \maincell{1.2716}{} & \maincell{1.463}{\pm0.258} & \maincell{1.250}{\pm0.216} & \maincell{1.356}{\pm0.188} \\
MDM~\citep{mdm} & \maincell{58.84}{\pm 0.09} & \maincell{73.51}{\pm 0.11} & \maincell{79.31}{\pm 0.06} & \maincell{0.9716}{} & \maincell{0.0897}{} & \maincell{1.3769}{} & \maincell{1.313}{\pm0.225} & \maincell{1.313}{\pm0.193} & \maincell{1.313}{\pm0.182} \\
MotionGPT3~\citep{motiongpt3} & \maincell{68.46}{\pm 0.12} & \maincell{81.29}{\pm 0.09} & \maincell{85.81}{\pm 0.08} & \maincell{0.8886}{} & \maincell{0.0628}{} & \maincell{1.3775}{} & \maincell{1.538}{\pm0.236} & \maincell{1.300}{\pm0.146} & \maincell{1.419}{\pm0.155} \\
TRACE (Ours) & \maincell{\mathbf{73.84}}{\pm 0.13} & \maincell{\mathbf{86.24}}{\pm 0.08} & \maincell{\mathbf{90.31}}{\pm 0.07} & \maincell{\mathbf{0.8165}}{} & \maincell{\mathbf{0.0434}}{} & \maincell{1.3766}{} & \maincell{\mathbf{2.338}}{\pm0.410} & \maincell{\mathbf{2.444}}{\pm0.407} & \maincell{\mathbf{2.391}}{\pm0.401} \\

%% file: fig/in_the_wild_look_left_right.tex
\begin{figure*}[t]
  \centering
  \includegraphics[width=\textwidth]{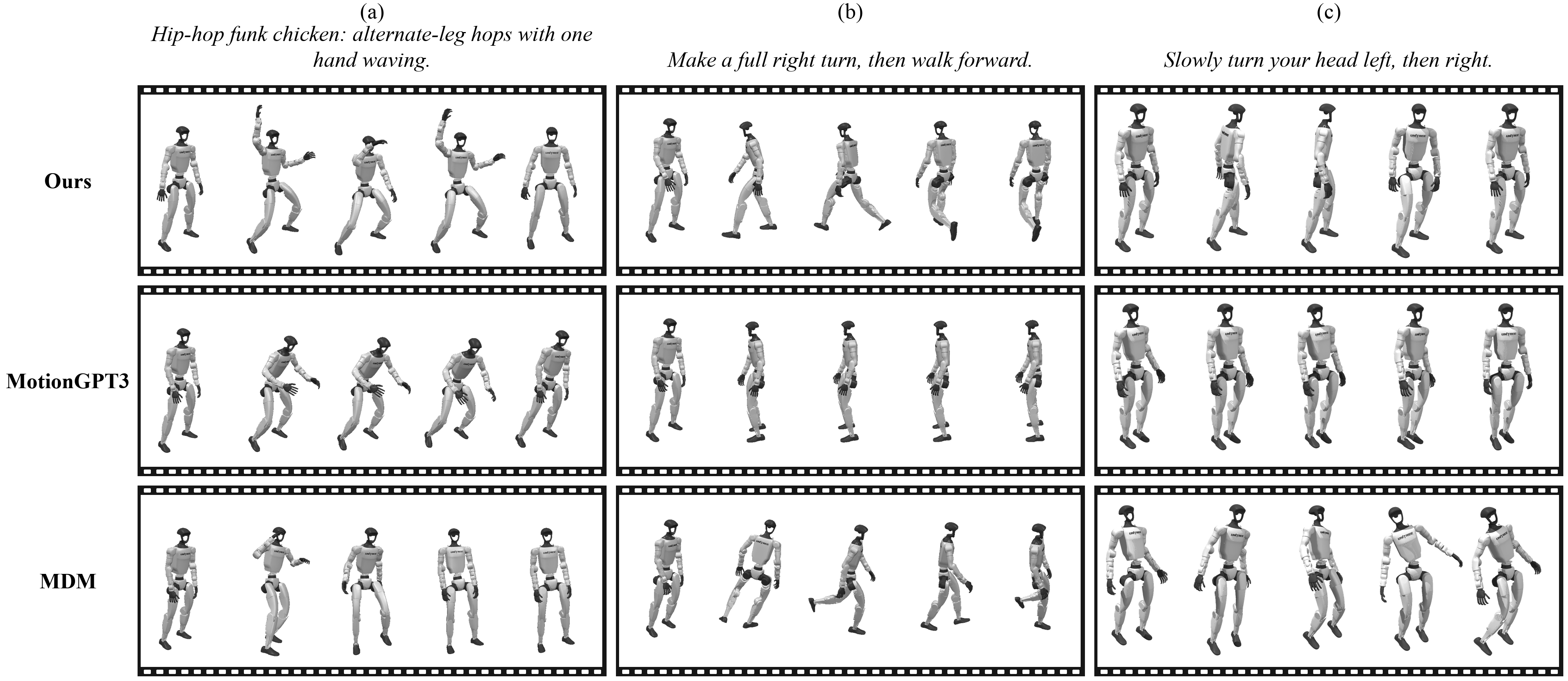}
  \caption{\textbf{Qualitative comparison of Ours, MotionGPT3, and MDM.}
  The third column illustrates generation from a free-form text prompt.}
  \label{fig:in_the_wild_look_left_right}
\end{figure*}

%% file: sec/contribution_ablations.tex
\subsection{Ablation Study}
\label{sec:ablation_studies}

\begin{figure}[!htbp]
\centering
\begin{minipage}[t]{0.485\linewidth}
\centering
\includegraphics[width=\linewidth]{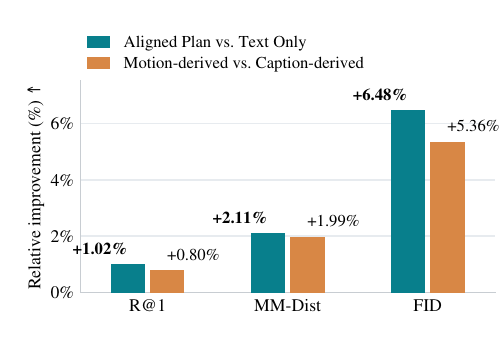}
\caption{\textbf{Motion-grounded Plan ablation.}
Teal compares Aligned Plan with Text Only; orange compares motion-derived
with caption-derived training supervision under a shared predicted test Plan.
Relative improvement is $100(v-b)/b$ for retrieval and $100(b-v)/b$ for
MM-Dist/FID. Positive is better, not percentage-point change.
Protocol details are in Appendix~\ref{sec:contact_plan_edit_protocol}.}
\label{fig:contact_plan_ablation}
\end{minipage}\hfill
\begin{minipage}[t]{0.485\linewidth}
\centering
\includegraphics[width=\linewidth]{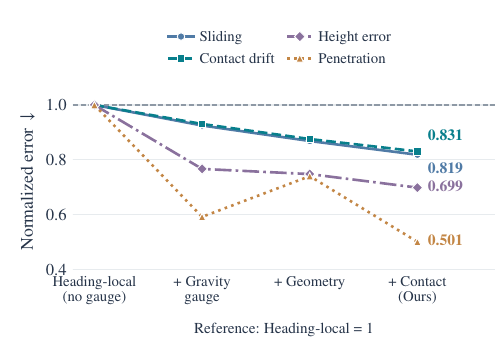}
\caption{\textbf{Reference-based kinematic evaluation.}
Each error is normalized by its Heading-local mean
(baseline 1; lower is better).
The first transition changes the state representation;
Geometry and Contact subsequently add objective bundles.
Contact-based measures use reference support intervals,
while penetration is evaluated over all valid frames.}
\label{fig:physical_design_controls}
\end{minipage}
\end{figure}

\paragraph{Motion-Grounded Plan Ablation.}
Matched controls support typed, time-aligned Plan conditioning and
supervision grounded in the paired motion
(Figure~\ref{fig:contact_plan_ablation}). Aligned Plan
uses the main-table model; Text Only retains the same raw-caption features,
branch capacity, duration predictor, initialization, objectives, and
continuation budget while zeroing the event inputs. A second comparison
replaces motion-derived auxiliary targets with caption-derived events
during training and retains the same predicted Plan at test time.

The retrieval and distributional improvements are consistent with two
roles of the Plan: temporal conditioning exposes when support and height
events occur, and motion-derived supervision aligns auxiliary targets
with the trajectory being reconstructed. A caption-derived Plan can
describe another valid realization. Test conditions are predicted from
text throughout. Appendix~\ref{sec:contact_plan_edit_protocol}
provides the Global Events control, absolute values, and full protocols.

\paragraph{Kinematic Design Ablation.}
\label{sec:kinematic_ablation}
The kinematic controls show complementary effects of integrated-trajectory
supervision and foot--ground constraints
(Figure~\ref{fig:physical_design_controls}). Heading-local,
Gravity-Gauged, Geometry, and Contact models share the training budget,
data, and event conditions. Gravity-Gauged reduces
contact-height error, consistent with predicting absolute height and tilt
directly. Geometry further improves trajectory-related diagnostics but
increases penetration; adding Contact reduces penetration and yields lower
errors than Heading-local across all reported measures.

Each plotted metric is normalized by its Heading-local mean.
Contact diagnostics use reference support
intervals, and penetration uses all valid frames; absolute values and
training details appear in Appendix~\ref{sec:physical_design_controls}.

%% file: sec/editing_experiments.tex
\section{Motion Editing Experiments}
\label{sec:editing_experiments}

We separately test automatic localization and multi-turn execution under
shared masks.

\subsection{Experimental Setup}
\label{sec:editing_setup}

\paragraph{Data and tasks.}
We use hand-height editing as a controlled test of source-relative request
execution. A requested height change defines a measurable FK goal while
allowing multiple valid joint trajectories, so evaluation can assess goal
achievement and preservation without requiring a unique target motion.
Within this protocol, the signed offset follows a fixed sine-squared profile
over the requested interval, using the benchmark's temporal convention.

We use 199 held-out BONES-SEED/G1 source motions with human-reviewed
synthetic English editing requests. Each initial request specifies a hand,
a signed 5\,cm height change, and an explicit time interval. We first
evaluate the predicted body--time mask without generating motion.
For sequential editing, each source yields three scripted four-turn
patterns: same-arm edits, alternating arms, and reversals, giving 597
chains.

\paragraph{Models and controls.}
For motion editing, we compare the complete TRACE editor
with local G1 adaptations of TMED~\citep{motionfix}
and SimMotionEdit~\citep{simmotionedit} under a shared
scope-compilation and reference-projection protocol.

\paragraph{Metrics.}
Mask localization uses body--time IoU against private reference masks.
All-turn success requires every request through a depth to meet change and
preservation criteria relative to its actual preceding output
(Appendix~\ref{sec:metric_definitions}).

\subsection{Scope-Construction Diagnostics}
\label{sec:automatic_mask_ablation}

\begin{figure}[htbp]
\centering
\begin{minipage}[t]{0.485\linewidth}
\centering
\includegraphics[width=\linewidth]{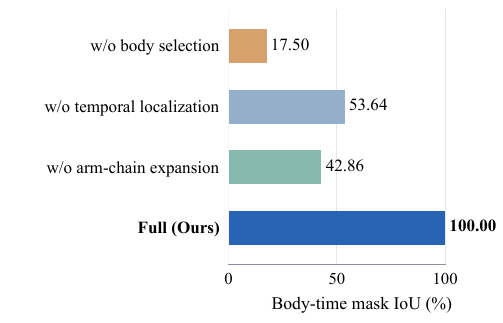}
\caption{\textbf{Scope diagnostics.} Body--time IoU on 199 reviewed requests;
parser outputs are fixed while compiler components are removed.}
\label{fig:automatic_mask_ablation}
\end{minipage}\hfill
\begin{minipage}[t]{0.485\linewidth}
\centering
\includegraphics[width=\linewidth]{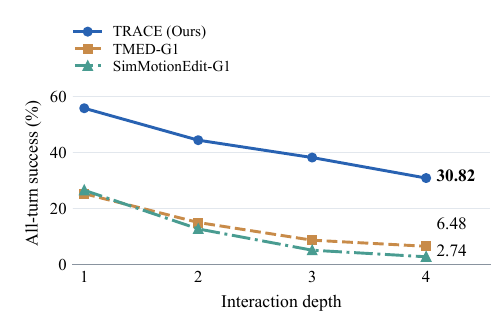}
\caption{\textbf{Editing under shared masks.} All-turn success over 597
scripted chains and three sampling runs; higher is better.}
\label{fig:multiround_shared_mask}
\end{minipage}
\end{figure}

\paragraph{Effect of scope-construction components.}
The full pipeline matches the prescribed masks on all 199 requests.
With parser outputs fixed, whole-body and full-duration masks
unnecessarily release unrelated coordinates or time intervals,
while wrist-only masks omit proximal coordinates of the prescribed
arm chain. These errors highlight the roles of body selection,
temporal localization, and kinematic-chain expansion in constructing
a compact editing scope, measured by agreement with prescribed masks
in Figure~\ref{fig:automatic_mask_ablation}.

\subsection{Multi-Turn Editing Analysis}
\label{sec:multiround_editing}

\paragraph{Multi-turn editing under shared preservation constraints.}
TRACE achieves the highest all-turn success at every evaluated depth.
At four turns, success reaches 30.82\%, compared with 6.48\% for
TMED-G1 and 2.74\% for SimMotionEdit-G1. All 597 chains remain in
the denominator, and each request is evaluated relative to the
method's preceding output. Shared reference projection gives all
methods zero protected-coordinate error, so the performance gap
reflects their ability to realize successive changes under the same
preservation constraint. Table~\ref{tab:current_edit_quality} provides
additional motion-quality diagnostics.

%% file: sec/5_conclusion.tex
\section{Conclusion and Limitations}
\label{sec:conclusion}

TRACE combines motion-grounded typed-event conditioning, gravity-referenced
diffusion, and source-relative editing to generate and revise Unitree G1
motion. Generation comparisons show strong semantic alignment and
perceptual quality; matched controls support temporal event conditioning,
motion-grounded supervision, and kinematic consistency. Under shared scopes
and reference projection, TRACE sustains successive hand-height changes
more reliably than the adapted learned editors. Typed motion quantities
thus connect language-conditioned generation with measurable revisions of
an evolving reference.

\paragraph{Scope and limitations.}
The system addresses text instructions and flat-ground kinematics without
scene perception. 
Retrieval and distributional metrics assess generated motions, not exact
adherence to each predicted event. Contact objectives encourage foot--ground
consistency without guaranteeing zero slip.

\newpage{}

%% file: sec/appendix.tex
\input{sec/real_robot_demos}
\clearpage

\clearpage
\section{Sparse Plan Design and Validation}
\label{sec:appendix_plan}

\subsection{Motion-Derived Sparse Plan Construction}
\label{sec:plan_annotation_rules}

Clean-motion support, root, and palm signals yield confidence-weighted
16-channel targets on the four-frame patch grid. The Plan is sparse in
vocabulary and dense in time; motion length is supplied separately.

\input{sec/plan_annotation_rules}
\input{sec/contact_plan_edit_protocol}

\input{sec/sparse_annotation_sensitivity}
\section{Kinematic Design and Objectives}
\label{sec:appendix_method}
\label{sec:appendix_kinematics}

\input{sec/gravity_gauge_representation}
\input{sec/contact_objective_details}
\input{sec/physical_design_protocol}

\section{Automatic Masking and Motion Editing}
\label{sec:appendix_editing}

\input{sec/language_frontend_fullbody_reproduction}
\input{sec/interaction_mdp_details}
\input{sec/editor_training_details}
\input{sec/mask_multiround_protocol}
\input{sec/edit_backend_validation}

\section{Data and Reproducibility Details}
\label{sec:appendix_reproducibility}

\input{sec/bones_seed_processing}
\input{sec/appendix_training_details}
\input{sec/inference_algorithm}
\input{sec/evaluation_metrics}


%% file: sec/real_robot_demos.tex
\clearpage
\section{Real-Robot Qualitative Demonstrations}
\label{sec:real_robot_demos}

\begin{figure}[!htbp]
\centering
{\small\textbf{(a) left--right looking}
``Can you slowly turn your head to the left and then to the right?''}\par\smallskip
\includegraphics[width=0.88\linewidth]{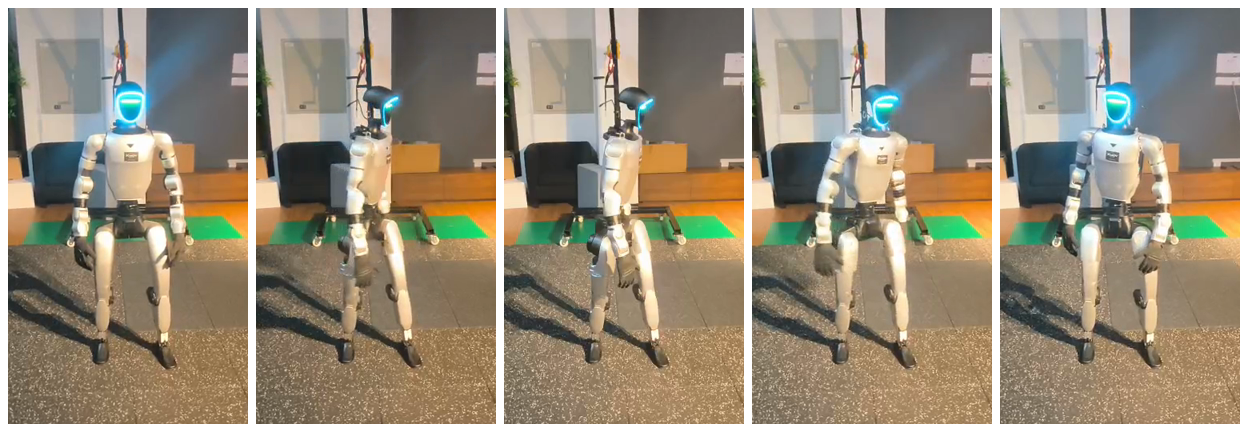}\par\medskip
{\small\textbf{(b) Hip-hop routine.}
``person quickly performs a hip-hop dance routine that includes a leg kick,
arm strike, and return''}\par\smallskip
\includegraphics[width=0.88\linewidth]{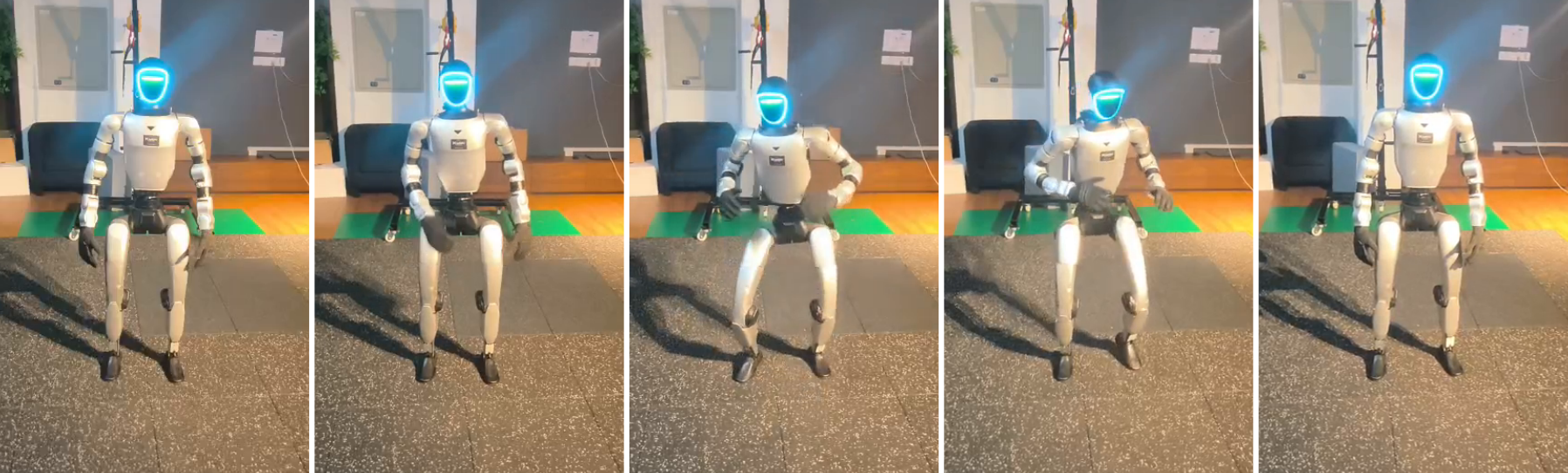}\par\medskip
{\small\textbf{(c) Weight-shifting groove.}
``grooving by shifting bodyweight from one leg to another, hands on hips
to hip hop music''}\par\smallskip
\includegraphics[width=0.88\linewidth]{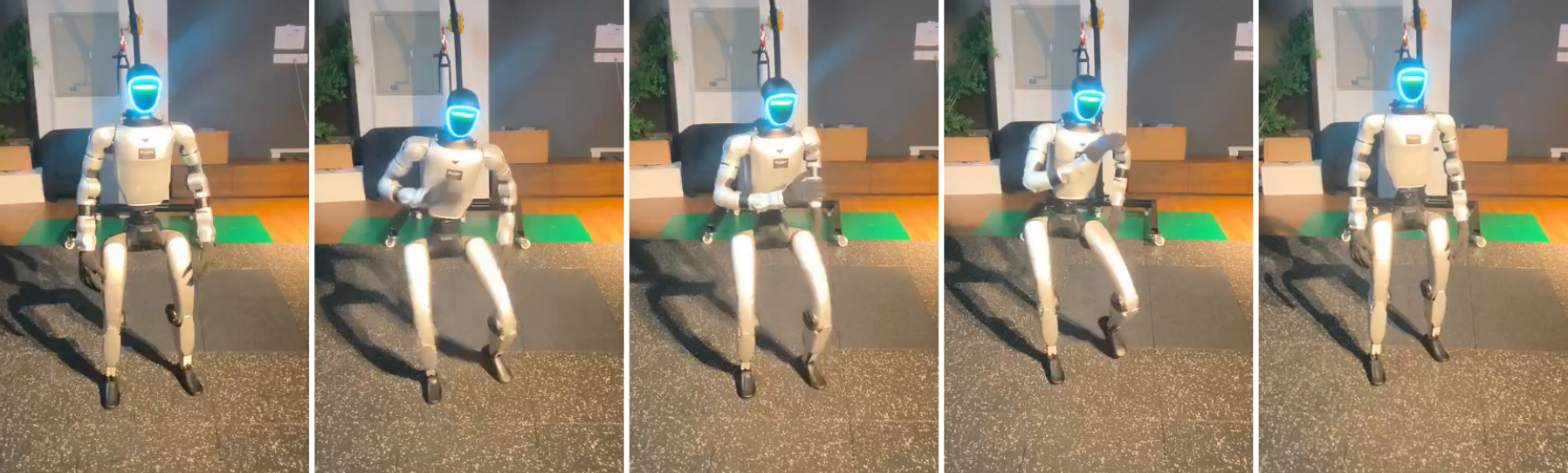}
\caption{\textbf{Real-robot demonstrations.} Panel (a) corresponds to the
in-the-wild request in Figure~\ref{fig:in_the_wild_look_left_right}.
Panels (b,c) sample 0/25/50/75/100\% of each recording. Fixed spatial crops
preserve poses and temporal order; clips are not time-synchronized.}
\label{fig:real_robot_demos}
\end{figure}

Figure~\ref{fig:real_robot_demos} presents real-robot demonstrations of
left--right looking, a hip-hop routine, and a weight-shifting groove.

%% file: sec/plan_annotation_rules.tex
\paragraph{Signals and coordinates.}
We use the clean G1 trajectory at 20 FPS, reconstructed from the
Gravity-Gauged state. Leg FK evaluates four sole points per foot; palm FK
follows the robot's hinge chains without clipping joint angles.
The ground $\bar z$ is the fifth-percentile order statistic of the pooled
minimum sole heights. Root height is measured above $\bar z$; hand height
is the vertical palm displacement relative to the pelvis in the
gravity-aligned frame.

\begin{table}[!htbp]
\centering\small
\caption{Sparse-event schema. Channels are zero-indexed.
The four categorical groups use softmax; the feet are independent
sigmoid probabilities. All distances are in meters.}
\label{tab:plan_annotation_thresholds}
\setlength{\tabcolsep}{4pt}
\begin{tabular}{p{.18\linewidth}p{.18\linewidth}p{.53\linewidth}}
\toprule
Channels & Quantity & Fixed states / rules\\
\midrule
0--1 & Left/right support & Enter at height $\leq0.035$ and speed
$\leq0.20$ m/s; leave at height $>0.055$ or speed $>0.30$ m/s.\\
2--4 & Root vertical speed & Descending / level / ascending,
split at $-0.08,+0.08$ m/s.\\
5--7 & Root height & Low / middle / high, split at $0.60,0.85$.\\
8--11 & Left hand height & H0 / H1 / H2 / H3, split at $0,0.25,0.45$
relative to pelvis.\\
12--15 & Right hand height & Same four bins as the left hand.\\
\bottomrule
\end{tabular}
\end{table}

\paragraph{Temporal annotation and uncertainty.}
Support uses hysteresis. Frames between the enter/leave thresholds are
uncertain, and contact runs shorter than three frames are confidence-masked.
Lack of stable support is not an airborne label.
For root/hand categories, we first apply a five-frame moving average with
edge padding; root vertical speed is the gradient of smoothed height times
20. Frames within 0.025 m of a height threshold or 0.02 m/s of a speed
threshold are uncertain. H3 is an operational height bin, despite its
implementation label \emph{above\_shoulder}; it is not measured relative
to an anatomical shoulder.

Each patch pools up to four valid frames. Its target is the histogram
over confident frames and its confidence is the fraction of confident
frames. No confident frame gives zero weight, not a negative event.
For each categorical group $g$ and class $j$,
\begin{equation}
 Y_{\ell,g,j}=
 \frac{\sum_{t\in B_\ell}w_{t,g}\mathbf1[y_{t,g}=j]}
      {\max(1,\sum_{t\in B_\ell}w_{t,g})},
 \qquad
 W_{\ell,g}=\frac{\sum_{t\in B_\ell}w_{t,g}}{|B_\ell|}.
\end{equation}
The same confident-frame pooling applies to each binary foot label.
The padded grid has 128 patches, of which $L=\lceil N/4\rceil$ are valid.
Patches use a fixed temporal grid.

\paragraph{Training and inference use.}
Training labels supervise the joint LLM planner, which is frozen during
diffusion training. Equation~\ref{eq:plan_mixture} specifies the confidence
fallback and 50/50 target/prediction mixture. Test conditions use only
captions and predicted durations; the main model's duration-planner prose
is not encoded as a condition. Training settings are detailed in
Appendix~\ref{sec:experiment_details}.

\input{tables/sparse_plan_example}

The example reproduces cached training targets.

%% file: tables/sparse_plan_example.tex
\begin{table}[!htbp]
\centering\small
\caption{Actual cached training Plan example. Caption: \emph{a simple jump in place}. $N=512$ frames (25.6 s), $L=128$ valid patches. The first non-mirrored jump UID in cache order is used. Patches show endpoints and first extrema of support, ascent, and confidence, selected solely from annotations, independently of model performance. Entries show support probability or categorical mode followed by confidence $W$.}
\label{tab:sparse_plan_example}
\setlength{\tabcolsep}{4pt}
\begin{tabular}{rcccccc}
\toprule
Patch & L foot & R foot & Root speed & Root height & L hand & R hand\\
\midrule
0 & 1.00/1.00 & 1.00/1.00 & level/1.00 & middle/1.00 & H0/1.00 & H0/1.00\\
4 & 0.00/0.50 & 0.00/0.50 & up/1.00 & high/0.75 & H0/1.00 & H0/1.00\\
87 & 0.67/0.75 & 0.50/1.00 & up/1.00 & middle/0.75 & H0/1.00 & H0/0.50\\
127 & 1.00/1.00 & 1.00/1.00 & level/1.00 & middle/1.00 & H0/1.00 & H0/1.00\\
\bottomrule
\end{tabular}
\end{table}
\noindent For example, patch 87 covers frames $[348,352)$. Its root-height target is $(0.00, 0.67, 0.33)$ (low, middle, high), not just the displayed mode. Its group confidence is $0.75$; the remaining weight falls back to the predicted distribution during target-conditioned training.

%% file: sec/contact_plan_edit_protocol.tex
\subsection{Main-Model Matched Plan Controls}
\label{sec:contact_plan_edit_protocol}

The Aligned Plan checkpoint is exactly the model in Table~\ref{tab:trace_main}.
All three controls retain identical residual capacity, raw-caption features,
32k-step EMA initialization, optimizer, losses, and 8k continuation budget.
They share the frozen joint event-and-duration LLM described
in Appendix~\ref{sec:experiment_details}.
Text Only zeros event inputs but retains trainable branch capacity.
Global Events averages every channel over valid patches and broadcasts
that mean at both training and inference. Aligned Plan preserves the
patchwise sequence. Both event-conditioned variants use
Equation~\ref{eq:plan_mixture}, isolating event content and timing under
the same training budget.

All arms use the same 13,302 test UIDs, EMA, DDIM-50, CFG 2.5, and batch
size 64.
Table~\ref{tab:main_plan_absolute} reports absolute semantic results.
Targets enter training supervision only, never test conditions.
These post-selection controls were not pre-registered.

\input{tables/main_plan_matched_absolute_20260925}

Aligned Plan improves the displayed semantic criteria over both controls,
consistent with patchwise events resolving temporal ambiguity left by
global averages: event prevalence alone omits when support and height
changes should occur.

We additionally isolate the source of Plan supervision by replacing the
motion-derived target in the 50\% target-conditioned training branch with
caption-derived events from a caption-only language model. The inference LLM
and its predicted test Plan remain fixed, as do initialization, raw-caption
features, residual capacity, 8k updates, objectives, and sampler. This is a
supervision-source control under a fixed inference planner.

\input{tables/plan_supervision_source_absolute_20260925}

Motion-derived supervision improves all reported retrieval and distributional
criteria. A caption admits several valid support, root, and hand trajectories;
caption-derived events can therefore disagree with the paired diffusion
target. Motion-derived events describe that target trajectory, supporting
the utility of realization-aligned supervision under the fixed inference
planner. Both variants predict the test Plan from text alone.

%% file: tables/main_plan_matched_absolute_20260925.tex
\begin{table}[htbp]
\centering\small
\caption{Matched LLM Plan controls on test. Aligned Plan is the main-table model. R@k is in percent.}
\label{tab:main_plan_absolute}
\begin{tabular}{lrrrrrr}
\toprule
Model & R@1$\uparrow$ & R@2$\uparrow$ & R@3$\uparrow$ & MM-Dist$\downarrow$ & FID$\downarrow$ & Diversity\\
\midrule
Text Only & 73.10 & 85.72 & 89.81 & 0.8341 & 0.0464 & 1.3817 \\
Global Events & 73.65 & 86.17 & 90.26 & 0.8310 & 0.0448 & 1.3809 \\
Aligned Plan (Ours) & \textbf{73.84} & \textbf{86.24} & \textbf{90.31} & \textbf{0.8165} & \textbf{0.0434} & 1.3766 \\
\bottomrule
\end{tabular}
\end{table}

%% file: tables/plan_supervision_source_absolute_20260925.tex
\begin{table}[htbp]
\centering\small
\caption{Matched Plan-supervision source control on test. Both rows use the
same predicted Plan at inference; only the training target source changes.
R@k is in percent.}
\label{tab:plan_supervision_source}
\begin{tabular}{lrrrrrr}
\toprule
Supervision & R@1$\uparrow$ & R@2$\uparrow$ & R@3$\uparrow$ & MM-Dist$\downarrow$ & FID$\downarrow$ & Diversity\\
\midrule
Caption-derived & 73.26 & 85.89 & 90.00 & 0.8331 & 0.0458 & 1.3817 \\
Motion-derived (Ours) & \textbf{73.84} & \textbf{86.24} & \textbf{90.31} & \textbf{0.8165} & \textbf{0.0434} & 1.3766 \\
\bottomrule
\end{tabular}
\end{table}

%% file: sec/gravity_gauge_representation.tex
\subsection{Gravity-Gauged State: Encoding and Decoding}
\label{sec:gravity_gauge_codec}

\paragraph{Coordinate convention and heading removal.}
Let $p_t\in\mathbb R^3$ and $R_t\in\mathrm{SO}(3)$ denote root position
and the body-to-world rotation at motion frame $t=0,\ldots,T-1$.
The world $z$ axis is vertical; the body's first axis is forward.
Positions are stored in centimeters and joint angles in degrees, while
the yaw angles below are in radians. Input root Euler angles are converted
from degrees to radians; $(\phi_t,\vartheta_t,\psi_t^{\mathrm{Euler}})$ form
$R_t=R_z(\psi_t^{\mathrm{Euler}})R_y(\vartheta_t)R_x(\phi_t)$.
The stored translation triplet $(p^{(0)},p^{(1)},p^{(2)})$ is reordered as
world $(x,y,z)=(p^{(2)},p^{(0)},p^{(1)})$ before encoding.
Write $e_x,e_y,e_z$ for the Cartesian unit vectors. Projecting
the root's forward axis onto the horizontal plane gives
\begin{equation}
 \begin{aligned}
 v_t&=(I-e_ze_z^\top)R_te_x, &
 f_t&=\begin{cases}
       v_t/\|v_t\|_2,&\|v_t\|_2>10^{-8},\\
       e_x,&\text{otherwise},
      \end{cases}\\
 H_t&=[f_t,\ e_z\!\times f_t,\ e_z]=R_z(\psi_t), &
 \psi_t&=\operatorname{atan2}(f_t^y,f_t^x).
 \end{aligned}
 \label{eq:gauge_heading_extraction}
\end{equation}
The yaw-free residual is $Q_t=H_t^\top R_t$. Its continuous six-coordinate
encoding stores the \emph{first two columns}, concatenated columnwise:
\begin{equation}
 \mathbf r_t^{\mathrm{tilt}}
   =\begin{bmatrix}Q_te_x\\Q_te_y\end{bmatrix}\in\mathbb R^6.
 \label{eq:gauge_tilt_encoding}
\end{equation}
Thus tilt encodes an absolute per-frame residual rotation.

\paragraph{Translation and yaw increments.}
For $t<T-1$, the remaining root coordinates are
\begin{equation}
 \begin{aligned}
 (\Delta p_t^x,\Delta p_t^y)
   &=\big[H_t^\top(p_{t+1}-p_t)\big]_{xy},
 &h_t&=p_t^z,\\
 \Delta\psi_t
   &=\operatorname{atan2}\!\left(\sin(\psi_{t+1}-\psi_t),
                                \cos(\psi_{t+1}-\psi_t)\right),
 &(s_t,c_t)&=(\sin\Delta\psi_t,\cos\Delta\psi_t).
 \end{aligned}
 \label{eq:gauge_increment_encoding}
\end{equation}
These per-frame displacements are integrated without a frame-rate factor.
Together with $h_t$, the six tilt coordinates
and 29 joint angles, they give the 40D state in
Equation~\ref{eq:gauge_root}. The encoder copies the penultimate planar
and yaw increments into the last frame; those last increments are never
integrated. For a single frame, planar increments are zero and
$(s_0,c_0)=(0,1)$.

\paragraph{Decoding and the initial gauge.}
Split a decoded tilt vector into $(a_{1,t},a_{2,t})\in\mathbb R^3\times
\mathbb R^3$. The implementation uses Gram--Schmidt normalization:
\begin{equation}
 \begin{aligned}
 \mathcal N_\epsilon(a)&=a/\max(\|a\|_2,\epsilon),\\
 b_{1,t}&=\mathcal N_\epsilon(a_{1,t}),\qquad
 b_{2,t}=\mathcal N_\epsilon\!\left(a_{2,t}
              -(b_{1,t}^\top a_{2,t})b_{1,t}\right),\\
 \widetilde Q_t&=[b_{1,t},\ b_{2,t},\ b_{1,t}\times b_{2,t}].
 \end{aligned}
 \label{eq:gauge_tilt_decoding}
\end{equation}
Here $\epsilon=10^{-6}$ in the differentiable FP32 Torch decoder and
$10^{-12}$ in the FP64 NumPy export decoder. The result is orthonormal
when both vectors have pre-normalization norms at least $\epsilon$;
zero or collinear pairs remain degenerate under this clamp.

For predicted $(s_t,c_t)$, recover
$\widetilde{\Delta\psi}_t=\operatorname{atan2}(s_t,c_t)$.
The Torch decoder substitutes $(0,1)$ if the pair norm is at most
$10^{-6}$; the NumPy export uses the pair directly.
Starting with $\widetilde\psi_0=0$ and $\widetilde p_0^{xy}=0$, decode
\begin{equation}
 \begin{aligned}
 \widetilde\psi_t&=\sum_{i=0}^{t-1}\widetilde{\Delta\psi}_i,
 &\widetilde H_t&=R_z(\widetilde\psi_t),\\
 \widetilde p_t^{xy}
   &=\sum_{i=0}^{t-1}\big[\widetilde H_i
                   (\Delta p_i^x,\Delta p_i^y,0)^\top\big]_{xy},
 &\widetilde p_t^z&=h_t,\\
 \widetilde R_t&=\widetilde H_t\widetilde Q_t.
 \end{aligned}
 \label{eq:gauge_full_decoding}
\end{equation}
The 29 joint angles are copied unchanged. Initial global planar position
and yaw are intentionally omitted: for an exactly encoded trajectory,
$\widetilde R_t=R_z(-\psi_0)R_t$. Restoring the original global frame
requires its initial planar translation and yaw as external metadata.
Height is retained directly.

\paragraph{Numerical input and masking.}
The denoiser operates on featurewise normalized states
$z_t=(x_t-\mu)/\sigma$, with training-only statistics and a standard
deviation floor of $10^{-4}$. Decoding first applies
$x_t=\sigma\odot z_t+\mu$. In a padded batch, each sequence has a valid
prefix; increments from padded frames are zeroed and padded outputs do
not enter any objective. The same canonical decoder is applied separately
to the prediction and clean target for the geometric losses below.

%% file: sec/contact_objective_details.tex
\subsection{Generation Objective Groups and Contact Implementation}
\label{sec:contact_implementation}

\paragraph{Objective groups.}
The compact objective in Section~\ref{sec:generation} collects the
component losses into three groups, with component-specific weights:
\begin{align}
 \mathcal L_{\mathrm{motion}}={}&\mathcal L_{\mathrm{rec}}+
 \lambda_v\mathcal L_{\mathrm{vel}}+\lambda_a\mathcal L_{\mathrm{acc}},\\
 \mathcal L_{\mathrm{geometry}}={}&
 \lambda_p\mathcal L_{\mathrm{int\text{-}pos}}+
 \lambda_\psi\mathcal L_{\mathrm{int\text{-}yaw}}+
 \lambda_g\mathcal L_{\mathrm{gravity}}+
 \lambda_u\mathcal L_{\mathrm{yaw\text{-}unit}},\\
 \mathcal L_{\mathrm{contact}}={}&
 \lambda_s\mathcal L_{\mathrm{slip}}+
 \lambda_b\mathcal L_{\mathrm{anchor}}+
 \lambda_h\mathcal L_{\mathrm{height}}+
 \lambda_f\mathcal L_{\mathrm{penetration}}.
 \label{eq:expanded_generation_groups}
\end{align}
Motion losses supervise normalized clean states and temporal derivatives;
geometry losses supervise integrated position/yaw, gravity, and yaw-unit
norm. Contact terms use decoded FK. Component activation is specified in
Appendix~\ref{sec:physical_design_controls}.

\paragraph{Geometric inputs and reduction.}
At a sampled diffusion noise level, the denoiser predicts a normalized
\emph{clean} sequence $\widehat z_{b,t}$, with clean target $z_{b,t}$.
Both are denormalized using the same training statistics and decoded by
Appendix~\ref{sec:gravity_gauge_codec}, yielding
$(\widehat p_{b,t},\widehat\psi_{b,t},\widehat R_{b,t})$ and
$(p_{b,t},\psi_{b,t},R_{b,t})$. Each trajectory integrates its own predicted
or target increments from zero initial planar position and yaw; there is
no additional rigid alignment or use of target headings to integrate
predicted translations. Geometry is computed from the unclipped one-step
clean prediction.

Let $m_{b,t}\in\{0,1\}$ be the valid-frame mask and
$V=\sum_{b,t}m_{b,t}>0$. Define body-up vectors
$\widehat u_{b,t}=\widehat R_{b,t}e_z$ and $u_{b,t}=R_{b,t}e_z$.
Let $(\widehat s_{b,t},\widehat c_{b,t})$ be the two denormalized yaw
coordinates \emph{before} the decoder's small-norm fallback. The implemented
geometry terms are
\begin{align}
 \mathcal L_{\mathrm{int\text{-}pos}}
 &=\frac{1}{3V}\sum_{b,t}m_{b,t}
       \left\|\frac{\widehat p_{b,t}-p_{b,t}}{s_p}\right\|_2^2,
       \qquad s_p=100,\label{eq:integrated_position_loss}\\
 \mathcal L_{\mathrm{int\text{-}yaw}}
 &=\frac{1}{V}\sum_{b,t}m_{b,t}
       \left[1-\cos(\widehat\psi_{b,t}-\psi_{b,t})\right],
       \label{eq:integrated_yaw_loss}\\
 \mathcal L_{\mathrm{gravity}}
 &=\frac{1}{3V}\sum_{b,t}m_{b,t}
       \|\widehat u_{b,t}-u_{b,t}\|_2^2,
       \label{eq:body_up_loss}\\
 \mathcal L_{\mathrm{yaw\text{-}unit}}
 &=\frac{1}{V}\sum_{b,t}m_{b,t}
       \left(\sqrt{\widehat s_{b,t}^{\,2}+\widehat c_{b,t}^{\,2}}-1\right)^2.
       \label{eq:yaw_unit_loss}
\end{align}
In Equation~\ref{eq:integrated_position_loss}, $p$ is the full three-coordinate
root position: its planar coordinates are integrated, while its height is
predicted directly. The factor $s_p=100$ converts centimeter-valued errors
to meters before squaring; division by $3V$ averages coordinates and
valid frames. Cumulative yaw enters the periodic cosine penalty directly
in radians.

The gravity term matches the \emph{target body-up direction}, preserving
demonstrated leaning. It uses the third column of the complete decoded
rotation, including heading composition.
The yaw-unit term penalizes $(\|v\|_2-1)^2$, the squared deviation of the
pair norm from one. The latter three terms are dimensionless. All sums
pool valid frames across the batch, so longer sequences contribute more
frames; there is no additional timestep-dependent reweighting. Empty
masks contribute zero, with implementation denominators clamped below by
one. Denormalization and geometric computations use FP32. The reported
geometry weights are
$(\lambda_p,\lambda_\psi,\lambda_g,\lambda_u)=(0.25,0.10,0.25,0.02)$.

\paragraph{Contact penalties.}
Let $p_{t,k}^f$ be sole point $k$ of foot $f$, $z_t^f=\min_k p_{t,k}^{f,z}$,
and $c_t^f$ the clean-motion support mask. Define uninterrupted-contact
weights $w_{t,\delta}^f=\prod_{j=0}^{\delta}c_{t+j}^f$. With masked means
over valid batch, time, feet, and applicable point/coordinate entries,
the implemented penalties are
\begin{align}
 \mathcal{L}_{\mathrm{slip}}&=
 \left\langle\rho_{\beta_v}\!\left(20(p_{t+1,k}^{f,xy}-p_{t,k}^{f,xy})\right)
 \right\rangle_{w_{t,1}^f},\\
 \mathcal{L}_{\mathrm{anchor}}&=\frac{1}{|\mathcal D|}\sum_{\delta\in\mathcal D}
 \left\langle\rho_{\beta_p}(p_{t+\delta,k}^{f,xy}-p_{t,k}^{f,xy})
 \right\rangle_{w_{t,\delta}^f},\\
 \mathcal{L}_{\mathrm{height}}&=
 \left\langle\rho_{\beta_h}(z_t^f-\bar z_{\mathrm{gt}})\right\rangle_{c_t^f},
 \label{eq:contact_losses}
\end{align}
where $\mathcal D=\{2,4,8\}$ excludes lags not shorter than the sequence.
Position and velocity use meters and meters per second, respectively.

The reported main checkpoint uses differentiable FK for both six-joint G1
legs. Each ankle-roll link carries four sole sample points, in meters:
$(-0.05,\pm0.025,-0.035)$ and $(0.12,\pm0.03,-0.035)$.
The ground estimate pools the clean minimum sole height of each foot over
valid frames and selects the fifth-percentile order statistic. A candidate
support frame has sole height at most 0.035\,m above this ground and foot-center
planar speed at most 0.20\,m/s. Singleton contacts are removed by requiring
an adjacent contact frame. All masks are computed from clean motion without
gradient propagation; generated motion is used only in the loss arguments.

Slip uses adjacent support frames; anchors require uninterrupted support
at lags 2/4/8 (0.1/0.2/0.4\,s), normalized separately. Height uses the
lowest sole point, not the ankle origin. Invalid frames are excluded;
empty masks contribute zero.
The elementwise robust penalty is
\begin{equation}
\rho_\beta(r)=\begin{cases}
r^2/(2\beta),& |r|<\beta,\\
|r|-\beta/2,&\text{otherwise}.
\end{cases}
\end{equation}
Its scales are $\beta_v=0.05$\,m/s for slip and
$\beta_p=\beta_h=0.01$\,m for anchor and height. All three loss weights are
0.05 in the reported checkpoint. Height supervision applies to support
frames; the penetration term below also covers swing frames.

\paragraph{Penetration and Plan-mask distinction.}
The current continuation and the final physical-control bundle also use
\begin{equation}
 \mathcal L_{\mathrm{penetration}}=
 \left\langle\max(0,\bar z_{\mathrm{gt}}-\widehat z_t^f)\right\rangle_{\mathrm{valid}},
\end{equation}
with weight one, zero tolerance, and FP32 contact computation.
This hinge averages over valid frames and feet, independently of support.

Loss masks use the clean-motion selector above, independently of predicted
Plan probabilities. Plan annotation instead uses hysteresis, confidence,
three-frame runs, and patch pooling (Appendix~\ref{sec:plan_annotation_rules}).

%% file: sec/physical_design_protocol.tex
\subsection{Motion-Consistency Ablation Protocol}
\label{sec:physical_design_controls}

All four controls use raw captions and the current sparse-event condition,
including the same 50/50 training mixture and frozen joint duration/event LLM.
They share training data, global batch 256, 32k updates from scratch,
learning rate $10^{-4}$, 500-step warm-up, and EMA 0.999.
Testing uses the same predicted events, DDIM-50, CFG 2.5, and 13,302
complete test motions.

All variants retain reconstruction, velocity, and acceleration losses.
Gravity-Gauged core has no geometry/contact objectives. Geometry adds
integrated-position, integrated-yaw, gravity, and yaw-unit losses with
weights $(0.25,0.10,0.25,0.02)$. Contact adds slip/anchor/height with
weight 0.05 each and penetration with weight one. The final step evaluates
Contact as a bundle.
The matched-budget representation comparison retains Heading-local and
Gravity-Gauged's native 38D/40D targets, normalization, and decoders.

\begin{table}[!htbp]
\centering\small
\caption{Absolute physical errors underlying Figure~\ref{fig:physical_design_controls}.
Mean $\pm$ SD over repeated sampling, in cm; lower is better. Sliding and
Drift measure planar endpoint displacement over reference contact runs,
using ankle and sole-point geometry, respectively.}
\label{tab:physical_absolute_controls}
\setlength{\tabcolsep}{3pt}
\begin{tabular}{lrrrr}
\toprule
Variant & Sliding & Drift & Height error & Penetration\\
\midrule
Heading-local & $28.152\pm0.090$ & $31.756\pm0.074$ & $3.385\pm0.018$ & $0.2659\pm0.0004$\\
Gravity-Gauged & $26.047\pm0.010$ & $29.558\pm0.023$ & $2.596\pm0.003$ & $0.1575\pm0.0021$\\
+ Geometry & $24.456\pm0.033$ & $27.817\pm0.055$ & $2.533\pm0.008$ & $0.1970\pm0.0017$\\
+ Contact & $23.047\pm0.062$ & $26.383\pm0.068$ & $2.366\pm0.007$ & $0.1333\pm0.0017$\\
\bottomrule
\end{tabular}
\end{table}

For variant $v$ and metric $m$, the plotted ordinate is
\begin{equation}
 \widetilde E_{v,m}=\frac{\operatorname{Mean}(E_{v,m})}
 {\operatorname{Mean}(E_{\mathrm{Heading},m})}.
 \label{eq:physical_plot_normalization}
\end{equation}
Each metric uses its own unrounded denominator and is reported separately.
Geometry increases penetration before Contact reduces it.

%% file: sec/language_frontend_fullbody_reproduction.tex
\subsection{Language Interfaces and Qualitative Whole-Body Example}
\label{sec:language_frontend_reproduction}

The two language adapters share the instruction-tuned backbone reported in
Section~\ref{sec:experimental_setup}, but are trained and serialized
independently because generation and editing expose different output schemas.
Both adapters are frozen when their respective diffusion models are trained or
sampled.

\paragraph{Adapter interfaces.}
The generation adapter receives a standalone instruction and emits a duration
plus a sequence of typed event intervals. Each interval records its start and
end, left/right support, root vertical state, root-height state, and left/right
hand-height states. Deterministic validation clips interval endpoints to the
predicted duration, resolves overlaps by output order, fills uncovered regions
with unknown states, and rasterizes the result onto the four-frame Plan grid.
The original instruction is encoded separately alongside the serialized Plan.

The editing adapter receives the follow-up instruction together with the duration and a source-only kinematic summary. It emits an attribute,
effector, signed change, and temporal interval. A deterministic compiler maps
this record to a body--time scope, a source-relative target when the attribute
has a metric interpretation, and revised source-event channels when the
request changes an event-level property. Invalid or incomplete records produce
a closed scope and leave the source unchanged.

\begin{table}[!htbp]
\centering\small
\caption{Serialized language interfaces. Both outputs are schema-validated
before being consumed by a motion model.}
\label{tab:language_adapter_interfaces}
\setlength{\tabcolsep}{4pt}
\begin{tabularx}{\linewidth}{lXX}
\toprule
Adapter & Supervised output & Saved inference artifact \\
\midrule
Generation & duration and typed support/root/hand intervals
& adapter weights, tokenizer/template identifier, schema, and event vocabulary \\
Editing & attribute, effector, signed change, and interval
& adapter weights, tokenizer/template identifier, schema, and compiler vocabulary \\
\bottomrule
\end{tabularx}
\end{table}

Training examples are stored as an instruction followed by one canonical JSON
object. Generation targets come from the motion-derived annotation in
Appendix~\ref{sec:plan_annotation_rules}; editing targets pair paraphrased
requests with typed deltas and compiler scopes. The reproducible checkpoint for
each adapter consists of its parameter-efficient weight update, the base-model
revision, prompt template, tokenizer files, output schema, and normalization or
vocabulary metadata. Decoding is constrained to one JSON object and followed
by the deterministic validator above. Each follow-up request is self-contained
and interpreted against the source state.

\paragraph{Whole-body refinement in Figure~\ref{fig:teaser}.}
Figure~\ref{fig:teaser} presents a jumping-jack refinement with a deeper
preparatory crouch and a higher jump. Its coordinated root and lower-body
changes are presented qualitatively, outside the hand-height benchmark
and its two-wrist target formulation and success criteria.

\paragraph{Evaluation scope.}
The reported editing benchmark starts from held-out G1 source motions so that
requested changes and protected coordinates can be measured against a common
source. After the first edit, each subsequent turn consumes the editor's own
preceding output, evaluating multi-turn accumulation. Quantitative results
therefore cover revision from held-out sources; Figure~\ref{fig:teaser}
illustrates the end-to-end generation-to-editing interface qualitatively.

%% file: sec/interaction_mdp_details.tex
\subsection{Markov Editing State and Relative Requests}
\label{sec:interaction_mdp_details}

For the supported self-contained requests, the state and transition follow
Equation~\ref{eq:interaction_factorization}: $X_t=(M_t,u_t)$ and
$X_{t+1}=(M_{t+1},u_{t+1})$, with external next utterance $u_{t+1}$.
The frozen parser emits a relative action, realized by supervised editing
and deterministic projection. The editor retains source duration and
derives its event conditions directly from the source motion.

\paragraph{Source-only automatic mask construction.}
Inputs are the request, frame count, and four-frame mean left/right wrist
heights relative to the pelvis, computed by source FK. The supervised
adapter predicts JSON fields for attribute, effector, signed offset, and
interval; the compiler validates them and resolves supported time syntax.
For hand $e^\star$, offset $d$, and interval $[a,b)$, compilation gives
\begin{align}
 S_{f,j}&=\mathbf1[j\in\mathcal J(e^\star)]\mathbf1[a\leq f<b],\\
 D_{f,h}&=\mathbf1[h=e^\star]\,d\sin^2\!\left(\frac{\pi(f-a)}{b-a-1}\right),
 \quad a\leq f<b,
 \label{eq:compiled_relative_request}
\end{align}
where $\mathcal J$ contains the seven arm coordinates; $D=0$ outside the
interval. The complement, including the selected arm outside its window,
is protected. This fixed envelope defines the synthetic task. The compiler
constructs the dense mask from typed fields using source-only inputs.

Offsets satisfy $0<|d|\leq0.10$\,m. Negated, ambiguous, or unsupported
expressions leave the source unchanged.

%% file: sec/editor_training_details.tex
\subsection{Reference-Conditioned Editor: Architecture and Training}
\label{sec:edit_backend_validation}

\paragraph{Training data.}
Retargeted official MotionFix pairs provide 5,387 training and 330 validation
pairs. Local supervision adds 3,690 pairs from 256 training sources, six
windows, both hands, signed 3/5/7\,cm offsets, and six instruction templates.
Source-only bounded IK constructs local training targets; rejected candidates
are recorded before evaluation. Selection uses 64 original-validation and
40 expanded-development cases. At inference, the learned editor produces
revisions through conditional denoising and reference projection, without per-request
IK optimization or post-hoc smoothing.

\paragraph{Architecture.}
The independent editor uses 40D states, four-frame patches, at most 104
frames, an eight-layer width-448 denoiser with eight heads, and a two-layer
source encoder. Text queries attend to source patches. Scope and two-wrist
offset patches have separate zero-initialized projections; offsets are
scaled by 5\,cm. Its weights and normalization are independent of the
main-table generator.

\begin{table}[!htbp]
\centering\small
\caption{TRACE editor training stages. Mixed batches contain equal numbers
of local and original MotionFix pairs.}
\label{tab:editor_training_stages}
\setlength{\tabcolsep}{4pt}
\begin{tabularx}{\linewidth}{lrX}
\toprule
Stage & Updates & Supervision / added condition\\
\midrule
Original editing & 8k & Original retargeted motion pairs\\
Mixed editing & 8k & Local/original source+scope initialization\\
Relative goal & 4k & Relative offsets and wrist goal loss\\
Event-aware & 4k & Source-derived event residual and consistency\\
\bottomrule
\end{tabularx}
\end{table}
Original-edit training uses batch 192 and learning rate $10^{-4}$.
Relative-goal and event-aware continuations use batch 128, learning rates
$5\times10^{-5}$ and $2\times10^{-5}$, AdamW, EMA 0.99, a 300-step
cosine schedule, and clean-state prediction. Selection maximizes mean joint success
on the two validation sets, breaking ties by height RMSE at steps 2k/4k.
Both TRACE continuations select step 4k.

\paragraph{Local objectives.}
Let $\mathcal L_{\mathrm{wrist}}$ denote wrist FK squared error normalized
by $(0.05\,\mathrm m)^2$. The local reconstruction and goal groups are
\begin{align}
 \mathcal L_{\mathrm{local}}&=\mathcal L_{\mathrm{rec,edit}}
       +0.2\mathcal L_{\mathrm{vel,edit}}+0.2\mathcal L_{\mathrm{wrist}},\\
 \mathcal L_{\mathrm{goal}}&=0.5\mathcal L_{\mathrm{vertical}}
                            +0.1\mathcal L_{\mathrm{horizontal}}.
\end{align}
The goal uses vertical and mean horizontal squared wrist errors, also
normalized by $(0.05\,\mathrm m)^2$, against the source trajectory plus
the public offset over editable entries.
Original-pair losses retain reconstruction,
velocity, and applicable auxiliary heads. Adapted baselines also receive
the local goal term.

\paragraph{Source-derived events.}
The final stage adds a zero-initialized MLP over 16 source-motion event
channels and four progress features. Only requested hand-height groups
are revised by the relative offset, keeping inference events source-derived.
The residual is attenuated during the last quarter of reverse
denoising. Confidence-weighted hand-category cross entropy has weight 0.05.
Current test cohorts and baseline checkpoints follow in
Appendix~\ref{sec:mask_multiround_protocol}; same-cohort quality diagnostics
are in Appendix~\ref{sec:current_edit_quality}.

%% file: sec/mask_multiround_protocol.tex
\subsection{Automatic Localization and Sequential-Editing Protocols}
\label{sec:mask_multiround_protocol}

\paragraph{Cohort and evaluation tasks.}
Of 200 frozen BONES-SEED/G1 candidates, 199 pass human review for
naturalness, scope, and feasibility, independently of model outputs.
Sources contain 64--104 frames; initial requests select one hand.
Mirrored duplicates, training-source groups, and exact source-array hashes
from earlier editing train/development sets are excluded.
Initial requests were reviewed; scripted chains were not independently
reviewed. We evaluate mask localization and shared-scope execution separately.

\paragraph{Mask-only structural interventions.}
We reuse the same frozen, fine-tuned instruction-parser predictions, whose
public inputs contain the instruction, length, and source summary.
Private reference masks are accessed only for scoring.
Removing body selection opens all 40 coordinates in the predicted
interval; removing temporal localization opens the selected arm throughout
the valid sequence. Removing arm-chain expansion retains only the three
wrist coordinates, compared with seven arm coordinates in the reference.
No model is retrained and no threshold is tuned in this diagnostic.

The whole-body overlap follows from the ratio of reference arm coordinates
to all state coordinates; wrist-only overlap follows from the retained
fraction of the arm chain. Full-duration overlap reflects the average
requested interval fraction. These geometric ratios explain agreement
with prescribed scopes in Figure~\ref{fig:automatic_mask_ablation}.

\paragraph{Four-turn construction and inference.}
Each source generates three fixed chains. Let $A$ be the original selected
hand, $B$ the other hand, and $d$ its signed 5\,cm offset. The same-arm
pattern is $(A{+}d,A{+}d,A{-}d,A{-}d)$; alternating arms uses
$(A{+}d,B{+}d,A{-}d,B{-}d)$; reversals use
$(A{+}d,A{-}d,A{+}d,A{-}d)$.
All methods receive the same explicit instructions, frame intervals,
deterministic compiler, and per-step reference projection. Each turn uses
the actual preceding generated motion, with no reset, teacher forcing,
early termination, or failure exclusion. The source duration is unchanged.
DDIM-50, CFG 1.5, EMA, batch 64, and paired sampling yield 21,492 outputs
across methods, turns, and chains.
Neither IK nor smoothing is applied to generated outputs.

\paragraph{Models and shared controls.}
Validation selects TRACE's event-aware step 4k and the mixed-training
goal-loss checkpoints at steps 4k/2k for TMED-G1/SimMotionEdit-G1, before
testing. TRACE retains relative offsets and source events; baselines retain
text/source interfaces. Scope and sampling are shared, but architectures,
prior training, and total FLOPs differ. The baselines are local G1 adaptations.

\paragraph{Aggregation, feasibility, and uncertainty.}
The primary denominator includes all 597 chains from 199 sources; no chain
is removed using solver-derived feasibility.
At depth $d$, all-turn success requires joint edit-and-preserve success
at every preceding turn, as defined in Appendix~\ref{sec:metric_definitions}.
We average repeated samples within each chain, then bootstrap 199 source
clusters while retaining all three patterns.
Four-turn success has 95\% source-cluster intervals of
$[26.07,35.62]$\% for TRACE, $[4.13,9.10]$\% for TMED-G1, and
$[1.23,4.47]$\% for SimMotionEdit-G1, quantifying source-sampling
uncertainty with trained models fixed.
Protected-coordinate error is zero for all methods under the shared
projection; success differences therefore concern requested changes.

%% file: sec/edit_backend_validation.tex
\subsection{Current-Cohort Editing Quality}
\label{sec:current_edit_quality}

TRACE improves four-turn success and the reported quality diagnostics
over the two learned adaptations.
We evaluate the same 199 sources and all 597 four-turn chains used in
Figure~\ref{fig:multiround_shared_mask}, aggregating paired sampling runs.
Each turn starts from the method's actual preceding output, without private
targets, feasibility filtering, or test-specific tuning.

\begin{table}[htbp]
\centering\small
\caption{Current-cohort editing comparison. Success requires all four
turns to pass. Quality diagnostics concern the fourth edit relative to
its incoming motion. Lower is better for all quality columns.}
\label{tab:current_edit_quality}
\setlength{\tabcolsep}{3pt}
\begin{tabular}{lrrrrr}
\toprule
Method & Success (\%) & XY RMSE (cm) & $\Delta v$ (deg/s) & $\Delta a$ (deg/s$^2$) & New limits (\%)\\
\midrule
SimMotionEdit-G1 & 2.74 & 2.238 & 34.273 & 602.284 & 0.033\\
TMED-G1 & 6.48 & 2.206 & 52.730 & 967.980 & 0.043\\
TRACE & 30.82 & 2.047 & 26.304 & 520.346 & 0.017\\
\bottomrule
\end{tabular}
\end{table}

XY RMSE measures unintended horizontal wrist displacement over the active
edit. Boundary diagnostics measure RMS velocity/acceleration changes
relative to the incoming motion on the edited arm. New-limit rate counts
editable coordinates that newly violate joint limits; pre-existing
violations are excluded from this count.

%% file: sec/bones_seed_processing.tex
\subsection{BONES-SEED Data Processing}
\label{sec:bones_seed_processing}

\paragraph{Source and scope.}
We use BONES-SEED's supplied G1 retargeting and captions~\citep{bonesseed}.
Its 142,220 metadata entries include original and mirrored motions.
We select a predominantly ground-supported subset and construct
study-specific grouped splits.

\paragraph{Selection and filtering.}
Requiring body-position \texttt{standing} and temporarily excluding mirrors
gives 61,292 originals. We reject malformed paths, Stunts, and metadata
indicating non-standing or externally supported motion. At an intermediate
50 FPS, a kinematic screen rejects clips whose two lowest foot-collision
points stay above world ground by 5\,cm for at least one second;
sampled rejection candidates are checked frame by frame. This leaves
58,574 originals.

A semantic filter removes 3,115 originals (1,969 keyword rules; 1,146
Qwen-Max classifications), leaving 55,459. It excludes difficult acrobatics
and externally supported actions, retaining walking, running, turns,
gestures, squats, daily dance, and small jumps. Persisted filtering decisions
are reused; semantic filtering is separate from deterministic Plan annotation.

\paragraph{Mirrors and captions.}
Restoring 55,423 supplied mirrors gives 110,882 clips; 36 originals remain
unpaired. No mirrors or captions are synthesized. We retain four natural,
one technical, and two short-description fields, plus filename-matched
temporal annotations. Mirror annotations come from their own records
without left/right substitution; clip-level captions may still be shared.
After whitespace/case-insensitive deduplication, the generator takes the
first nonempty caption in this order; semantic evaluation uses the
corresponding first natural description.

\paragraph{Grouped split and leakage checks.}
We connect records sharing a motion family, its mirror-canonical family,
an original/mirror pair, or a normalized exact caption. Caption matching
uses all available English alternatives and an existing locally translated
Chinese copy of the same metadata. Text normalization uses
Unicode normalization, case folding, and punctuation/whitespace cleanup.
Connected components are kept intact when assigning train/validation/test
with seed 20260822 and target proportions 76/12/12, balancing clips and
frames. Adding mirrors rebuilds the assignment and produces 2,557 final
components. The recorded audit finds zero cross-split overlap for paths,
exact and mirror-canonical families, caption keys, components, and mirror
pairs. This is not an actor-disjoint split or a guarantee that every
semantic paraphrase is disjoint; generic temporal annotation phrases are
not grouping keys.

\begin{table}[htbp]
\centering
\caption{Processed BONES-SEED counts. Full clips, packed windows, and
eligible model examples are different units. Eligible windows have at
least ten frames; durations count full clips at 20 FPS.}
\label{tab:bones_seed_processing}
\small
\setlength{\tabcolsep}{6pt}
\begin{tabular}{lrrrrr}
\toprule
Split & Originals & Mirrors & Full clips & Packed / eligible windows & Hours \\
\midrule
Train & 42,156 & 42,131 & 84,287 & 86,451 / 86,385 & 170.54 \\
Validation & 6,650 & 6,643 & 13,293 & 13,803 / 13,799 & 26.95 \\
Test & 6,653 & 6,649 & 13,302 & 13,722 / 13,706 & 26.91 \\
\bottomrule
\end{tabular}
\end{table}

\paragraph{Temporal and geometric representation.}
The final 20 FPS trajectories are obtained directly from the 120 FPS
source: output frame $k$ copies source frame $6k$, without interpolation,
smoothing, or IK. Frame counts and durations are recomputed at the target
rate. Parsing retains root translation/orientation and the 29 G1 joint
angles; unused non-root translation fields do not enter the model.
We then convert to the 40D Gravity-Gauged state defined in
Section~\ref{sec:generation}: planar increments, absolute root height,
yaw increments, a six-coordinate yaw-free tilt rotation, and joint angles.
Appendix~\ref{sec:gravity_gauge_codec} defines heading removal, columnwise
tilt encoding, canonical root reconstruction, and terminal-frame handling.
Means and standard
deviations are estimated from training frames only, with a standard
deviation floor of $10^{-4}$; validation and test reuse these statistics.

\paragraph{Windowing and Plan supervision.}
Long clips are split into consecutive, nonoverlapping windows of at most
512 frames. Original/mirror pairs share window boundaries, and each
window retains its parent motion ID, source offsets, and ordering.
Temporal annotations are clipped and rebased to each window. Windows
shorter than ten frames are excluded from the model's input
manifests (66/4/16 in train/validation/test); padded frames are masked out
of losses. Four-frame patches are formed only after these operations.
The sparse-event Plan is derived from clean G1 motion through FK,
thresholding, confidence masking, and patch aggregation, independently of
the provider's prose phase labels.
Appendix~\ref{sec:plan_annotation_rules} specifies every event channel
and its thresholds. Training-motion labels supervise the joint LLM planner;
test conditioning uses predicted events and durations, never target
event labels.

\paragraph{Full-motion evaluation.}
The 13,706 eligible test windows are generated, decoded, and reassembled
in parent-window order into exactly 13,302 full-motion IDs. The 
variable-length assembly concatenates heading-local root increments and
joint states; it does not pad or truncate predictions to GT lengths and
does not fill excluded short tails with GT frames. Each parent is scored
once. Source-derived segmentation is retained, so the protocol predicts
window durations but is not an unconstrained full-clip duration benchmark.

%% file: sec/appendix_training_details.tex
\subsection{Training and Evaluation Configuration}
\label{sec:experiment_details}

\paragraph{Generation backbone and continuation.}
The frozen encoder in Section~\ref{sec:experimental_setup} supplies a
2,560-dimensional caption vector. The generator has 32,508,520 parameters,
including residual branches but excluding upstream models, and feed-forward
width 1,792. Training predicts clean states with a 1,000-step cosine schedule.

\begin{table}[!htbp]
\centering\small
\caption{Generation training schedule. Continuations start from the same
raw-text EMA initialization and retain equal residual capacity. Physical
from-scratch controls are specified in Appendix~\ref{sec:physical_design_controls}.}
\label{tab:generation_training_schedule}
\begin{tabular}{lrr}
\toprule
Setting & Raw-text backbone & Sparse-event continuation\\
\midrule
Updates & 32k & 8k\\
Global batch & 256 & 128\\
Learning rate & $10^{-4}$ & $2\times10^{-5}$\\
\bottomrule
\end{tabular}
\end{table}

Continuation uses 250-step warm-up and EMA 0.999. Contact losses use FP32,
with weights in Appendix~\ref{sec:contact_implementation}. The event residual
is scaled by $\min(1,k/[0.25(K-1)])$, $K=1000$. Main-table inference uses
EMA, DDIM-50, CFG 2.5, and batch 64.

\paragraph{Joint LLM planner.}
A supervised language-model adapter predicts duration and typed support,
root, and hand events in a structured Plan. The event intervals are
rasterized onto the four-frame patch grid; no separate support or
root/hand prediction network is used in this configuration.
The LLM is frozen for diffusion training and inference.
The generator uses all 86,385 training segments with the joint 50/50
condition mixture in Equation~\ref{eq:plan_mixture}; event extraction is
defined once in Appendix~\ref{sec:plan_annotation_rules}.
Most predicted training conditions are in-sample, without out-of-fold
separation. Test inputs contain no target events or durations.
The Plan controls in Appendix~\ref{sec:contact_plan_edit_protocol} reuse
the same joint LLM frontend. Invalid Plans remain in the evaluation with
target-free fallback: unknown support values are 0.5 and unknown
categorical groups use uniform probabilities.

\paragraph{Retrained baselines.}
MDM, MoMask, and MotionGPT3 consume raw captions and predict heading-local
38D G1 motion. Their budgets are 64k steps (MDM), 17k VQ-autoencoder plus
50k base/residual Transformer steps (MoMask), and 50k motion-VAE plus 50k
language-model steps with pretrained language initialization (MotionGPT3).
ECHO, TextOP, and TeXedo retain official weights and generation
settings.

%% file: sec/inference_algorithm.tex
\subsection{Inference Procedure}
\label{sec:inference}

Algorithm~\ref{alg:system} distinguishes the main generation network
$F_\theta$ from the independent editor $E_\eta$. Editing uses source
features in addition to reference projection; generation predicts sparse
events from text.

\begin{algorithm}[!htbp]
\caption{Sparse generation and independent relative editing}
\label{alg:system}
\begin{algorithmic}[1]
\REQUIRE instruction $u$, optional source motion $M$
\STATE $e\gets f(u)$
\IF{$M=\varnothing$}
 \STATE $(N,C)\gets\textsc{Rasterize}(\textsc{LLMPlan}(u))$; $x_K\sim\mathcal N(0,I)$
\ELSE
 \STATE $N\gets\operatorname{length}(M)$; $\delta\gets\textsc{ParseLLM}(u,N,s(M))$
 \STATE $(S,D)\gets\textsc{ValidateAndCompile}(\delta,u,N)$
 \IF{parsing or compilation fails}
  \RETURN $M$
 \ENDIF
 \STATE $m\gets\operatorname{Enc}(M)$
 \STATE $x_K\gets S\odot\epsilon+(1-S)\odot q_K(m)$
\ENDIF
\FOR{each reverse DDIM timestep $k$}
 \IF{$M=\varnothing$}
  \STATE $\widehat x_0\gets F_\theta(x_k,k,e,C,N)$ with CFG
 \ELSE
  \STATE $\widehat x_0\gets E_\eta(x_k,k,e,m,S,D)$ with CFG
 \ENDIF
 \STATE $\widetilde x_{k-1}\gets\textsc{DDIM}(x_k,\widehat x_0,k)$
 \IF{$M\neq\varnothing$}
  \STATE $x_{k-1}\gets S\odot\widetilde x_{k-1}+(1-S)\odot q_{k-1}(m)$
 \ELSE
  \STATE $x_{k-1}\gets\widetilde x_{k-1}$
 \ENDIF
\ENDFOR
\RETURN $\textsc{Decode}(x_0)$
\end{algorithmic}
\end{algorithm}

%% file: sec/evaluation_metrics.tex
\subsection{Metric Definitions and Aggregation}
\label{sec:metric_definitions}

\subsubsection{Semantic and Distributional Metrics}

All main-table methods use one frozen G1 evaluator and 13,302 parent IDs.
The text branch uses GloVe/POS features and a bidirectional GRU; the motion
branch uses normalized 38D states, temporal convolutions, and a bidirectional
GRU. Both yield unit-normalized 512D embeddings. Training uses a symmetric
contrastive objective on the training split, with no generated-test updates.
Let $e_i$, $z_i$, and $z_i^{\mathrm{gt}}$ be paired text, generated-motion,
and target-motion embeddings. Distances are Euclidean.

\paragraph{R-Precision (R@1/R@2/R@3, higher is better).}
For every method, retrieval is motion-to-text: $z_i$ ranks its caption
among 32 candidates, including 31 others sampled without replacement.
For $k\in\{1,2,3\}$,
\begin{equation}
 \mathrm{R@}k=\frac{100}{N}\sum_{i=1}^{N}
 \mathbf{1}\!\left[\operatorname{rank}(e_i\mid z_i)\leq k\right].
 \label{eq:metric_r_precision}
\end{equation}
We average 20 candidate samplings to measure paired-caption alignment.

\paragraph{Multimodal distance (MM-Dist, lower is better).}
The mean paired distance $N^{-1}\sum_i\|z_i-e_i\|_2$ measures text--motion
alignment, not within-caption diversity.

\paragraph{Fr\'echet Inception Distance (FID, lower is better).}
FID compares the full generated/GT embedding distributions, with means
$\mu_g,\mu_r$ and covariances $\Sigma_g,\Sigma_r$:
\begin{equation}
 \mathrm{FID}=\|\mu_g-\mu_r\|_2^2+
 \operatorname{tr}\!\left(\Sigma_g+\Sigma_r
             -2(\Sigma_g\Sigma_r)^{1/2}\right).
 \label{eq:metric_fid}
\end{equation}
Features come from the G1 encoder; FID is computed once per complete set.

\paragraph{Diversity (reference diagnostic).}
We average $\|z_i-z_j\|_2$ over 300 distinct-motion pairs in each of 20
repetitions and compare with GT. Higher is not necessarily better.
Multimodality, by contrast, requires multiple generations per fixed caption.

\subsubsection{Kinematic Metrics}

G1 FK is evaluated on the common predicted/GT prefix, without time warping.
Within-clip scores are averaged equally across clips, then across repeated
sampling runs. GT contacts are scoring masks only. These diagnostics
measure adherence to reference contact timing as well as kinematic
consistency: a valid generated swing during GT support can increase the
error. They do not isolate foot slip under the generated motion's own
contacts.

\paragraph{Foot Sliding (cm, lower is better).}
Ground is the fifth percentile of the framewise lower GT ankle height
over the full target clip. Support requires height above ground at most
3.5\,cm and planar speed at most 0.20\,m/s; singleton contacts are removed.
Sliding averages generated-ankle planar endpoint displacement over contact
runs of both feet; endpoint scores can miss within-run excursions.

\paragraph{Contact Drift (cm, lower is better).}
Using four sole corners per foot, ground is the fifth percentile of pooled
GT lowest-sole heights on the evaluated prefix. Support requires lowest-sole
height within 3.5\,cm above ground and sole-centroid planar speed at most
0.20\,m/s, excluding singletons. Drift averages generated corners' planar
endpoint displacements over corners, runs, and feet. Its geometry and mask
differ from Sliding. Clips without valid runs contribute zero to both metrics.

\paragraph{Contact Height Error (cm, lower is better).}
For generated lowest-sole height $\hat h_{t,f}$ and the same sole ground
$g$, average $|\hat h_{t,f}-g|$ over GT-contact frames and feet.
Clips without selected contacts contribute zero.

\paragraph{Penetration (cm, lower is better).}
Average $\max(g-\hat h_{t,f},0)$ over \emph{all} valid frames and feet.
This measures depth, not the fraction of penetrating frames. All four
errors are converted to centimeters. Figure~\ref{fig:physical_design_controls}
normalizes them separately by Heading-local means
(Equation~\ref{eq:physical_plot_normalization}); no aggregate score is formed.

\subsubsection{Editing and Human Ratings}

\paragraph{Relative edit success.}
Let $z_f^{\mathrm{src}}$ and $z_f^{\mathrm{edit}}$ be the requested wrist's
world-space FK height, $\sigma\in\{-1,1\}$ the requested direction, and
$w_f$ the fixed sine-squared envelope. For the test request of 5\,cm,
success requires median signed gain
$\operatorname{median}_{w_f\geq0.8}\sigma(z_f^{\mathrm{edit}}-z_f^{\mathrm{src}})
\geq3$\,cm and height RMSE at most 2.5\,cm over the complete editable
interval, relative to the profile $\sigma\,5w_f$\,cm. The score evaluates
the requested relative height profile.

\paragraph{Preservation and joint success.}
Scoring uses the private reference mask, never a method's predicted mask.
Preservation requires at most 1\% of protected joint/frame coordinates to
change by more than $1^\circ$, protected FK position error at the 95th
percentile at most 1\,cm, maximum root displacement at most 1\,cm and
rotation change at most $1^\circ$, and unchanged duration.
Joint success is the conjunction of edit and preservation success.
All compared editors share projection; their protected joint MAE is zero.
All requests and parse failures remain in the denominator.
Same-cohort quality diagnostics are reported in
Appendix~\ref{sec:current_edit_quality}.

\paragraph{Automatic body--time mask overlap.}
Let $\widehat S_i$ and $S_i^\star$ be the predicted and private reference
sets of editable frame--coordinate pairs within valid frames.
Figure~\ref{fig:automatic_mask_ablation} reports coordinate-level macro IoU:
\begin{equation}
 \mathrm{MaskIoU}=\frac{100}{N}\sum_i
 \frac{|\widehat S_i\cap S_i^\star|}{|\widehat S_i\cup S_i^\star|}.
 \label{eq:mask_iou_metric}
\end{equation}
An empty prediction against a nonempty reference scores zero.

\paragraph{All-turn editing success.}
Let $J_{i,r,t}$ indicate joint edit-and-preserve success at turn $t$
of chain $i$ in sampling replicate $r$. Each request is scored relative
to the actual preceding output using the unchanged thresholds above.
For $N=597$ chains and $R=3$ replicates, the curve in
Figure~\ref{fig:multiround_shared_mask} is
\begin{equation}
 \mathrm{AllTurn}(d)=\frac{100}{NR}
    \sum_{i=1}^{N}\sum_{r=1}^{R}\prod_{t=1}^{d}J_{i,r,t}.
 \label{eq:all_turn_success}
\end{equation}
An earlier failure persists at greater depths; no failed chains are
removed. Source-cluster uncertainty follows
Appendix~\ref{sec:mask_multiround_protocol}.

\paragraph{Auxiliary whole-body and boundary diagnostics.}
World-space FK MPJPE averages all 29 actuated joint-body positions against
the private IK witness without alignment, retaining coincident origins.
The witness provides one reference realization. Boundary error measures
edited-arm velocity change from source at both interval edges (20 FPS finite
differences, deg/s). These diagnostics supplement request satisfaction.

\paragraph{User Study.}
The interface hides method names and identifies videos only by letters
A--H; the letter-to-method mapping is stored separately from the survey
page. For each prompt, method assignments to letters are randomly permuted
once with a fixed seed; the resulting order is shared across participants.
Videos appear together
with the prompt in a grid, using a common G1 rendering setup, background,
and camera configuration, without method-name overlays. Participants can
pause and replay the clips and rate every video on two separate criteria:
motion Quality and Text match.

The displayed three-point anchors are: \textbf{1 (poor)}, clearly
unnatural or discontinuous motion with severe visible artifacts, or a
mismatch with the main action or intent; \textbf{2 (fair)}, broadly
plausible motion with local coordination defects or mild artifacts, or
a broadly correct action with inaccurate details, direction, amplitude,
or timing; and \textbf{3 (good)}, natural, continuous, coordinated motion
with few visible artifacts, or accurate expression of the principal action,
direction, and intent. The two criteria are scored independently. Score definitions
remain visible during rating, and all required ratings must be supplied
to complete an item. Overall is the equal-weight mean of the two criteria.
We first average each participant's ratings for a method, then report the
mean and sample SD across participants.


